\documentclass[]{interact}

\usepackage{amssymb,amsmath}
\usepackage{indentfirst}
\usepackage{graphicx}
\usepackage{subcaption}
\DeclareMathOperator*{\argmin}{argmin}
\usepackage{algorithm}
\usepackage{algorithmicx}
\usepackage{algpseudocode}
\usepackage{multirow}
\usepackage{float}
\usepackage{url}
\usepackage{threeparttable}
\usepackage{epstopdf}
\usepackage{appendix}
\usepackage[longnamesfirst,sort]{natbib}
\bibpunct[, ]{(}{)}{;}{a}{,}{,}

\theoremstyle{plain}

\theoremstyle{definition}

\theoremstyle{remark}

\title{Gradient Boosting Survival Tree with Applications in Credit Scoring}
\author{Miaojun Bai, Yan Zheng and Yun Shen\\
{\small baimiaojun-jk, zhengyan, shenyun-jk@360jinrong.net}\\ 360 Finance Inc.}

\date{}

\begin{document}

\maketitle

\begin{abstract}
 Credit scoring plays a vital role in the field of consumer finance. Survival analysis provides an advanced solution to the credit-scoring problem by quantifying the probability of survival time. In order to deal with highly heterogeneous industrial data collected in Chinese market of consumer finance, we propose a nonparametric ensemble tree model called gradient boosting survival tree (GBST) that extends the survival tree models with a gradient boosting algorithm. The survival tree ensemble is learned by minimizing the negative log-likelihood in an additive manner. 
 The proposed model optimizes the survival probability simultaneously for each time period, which can reduce the overall error significantly. 
 Finally, as a test of the applicability, we apply the GBST model to quantify the credit risk with large-scale real market datasets. The results show that the GBST model outperforms the existing survival models measured by the concordance index (C-index), Kolmogorov-Smirnov (KS) index, as well as by the area under the receiver operating characteristic curve (AUC) of each time period.
\end{abstract}


\section{Introduction}

Chinese consumer finance market, with the rapid development of internet finance and e-commerce, has entered a period of market eruption in recent years. The consumer credit hit 8.45 trillion RMB, the historical high in October 2018, according to a research report released by Tsinghua University\footnote{Chinese consumer credit market research 2018 (in Chinese), Institute for Chinese Economic Practice and Thinking, Tsinghua University, \url{https://att.dahecube.com/4a5ab7de-1aa3-4754-a4d7-6e707d943d69}.}. However, only approximately one third of Chinese consumers have credit ratings, in spite of the rising needs for financing, and therefore, the lack of credit infrastructure has made a critical challenge to Chinese credit providers and it calls for more advanced credit scoring techniques. 

Credit scoring has been a commonly used risk measurement tool for decades (see \cite{thomas2002credit} for a comprehensive introduction) and it has already been widely adopted by credit agencies all over the world.  Credit scoring is traditionally solved by classification methods, e.g., logistic regression, decision tree, and more recently, gradient boosting decision tree (GBDT, \citealp{friedman2002stochastic}) and other boosting models, e.g., XGBoost (XGB, \citealp{chen2016xgboost}). 

Survival analysis is a statistical method to deal with the analysis of time-to-event data, where the time to the occurrence of an event is of particular interest. In the context of credit scoring, the event is usually defined as ``default''. Therefore, the primary advantage of applying survival analysis is that it can not only model whether to default, but also predict the probability of time when default occurs. The idea to use survival analysis in the field of credit scoring was first introduced by \cite{narain1992survival2}. Since 2000, several studies (\citealp{cao2009modelling, dirick2017time} and references therein) have applied survival analysis to quantify the credit risk and the results show a better performance on real credit data, which are collected by banks in developed countries like the UK and Belgium \citep{dirick2017time}, than the traditional credit scoring methods. In contrast to the developed countries where credit data are mostly provided by credit bureaus, the data collected for credit assessment in Chinese market are usually obtained from the Internet, which makes them high-dimensional, noisy, sparse and extremely heterogeneous. 

The most widely applied model in survival analysis is the Cox proportional hazards model \citep{blakely1972evidence} with (generalized) linear functions \citep{cao2009modelling} or with splines \citep{dirick2017time}. To relax the linear assumption imposed by Cox model, several extensions were developed. Among them, a boosting approach (CoxBoost) was introduced by \cite{binder2008allowing}, while a gradient boosting algorithm called GBMCI was developed by \cite{chen2013gradient} to optimize the concordance index (C-index, \citealp{harrell1982evaluating}) directly. Both algorithms are using Cox as base models. 

An alternative category of survival analysis methods is based on trees, e.g., survival trees \citep{gordon1985tree} and random survival forests (RSF, \citealp{ishwaran2008random}) were proposed to quantify survival probabilities in medical studies. \cite{wright2017unbiased} proposed the conditional inference survival forest model (CIF) to correct the bias in the RSF model, and empirical studies \citep{nasejje2017comparison} showed that CIF is superior to RSF model on some medical datasets.

Recently several models originated from the machine learning community are adapted to survival analysis. For instance, \cite{fernandez2016gaussian} introduced a semi-parametric Bayesian survival model based on Gaussian processes, while an approach DeepHit based on deep neural networks was introduced by \cite{lee2018deephit} to learn the distribution of survival times. The former model relies on the assumption that the latent stochastic process follows a Gaussian process which is usually not true for real credit data. The problem of applying the latter approach to credit scoring is that deep neural networks perform in general worse on tabular data (like the credit data) than tree-based ensemble methods, 
unless the network architecture is specially designed \citep{2021tabnet, Popov2020Neural}. DeepHit introduced by \cite{lee2018deephit} (see also \citealp{blumenstock2020deep}) is based on the fully-connected type of network, which may not be the best architecture for heterogeneous credit data.


On the one hand, to our best knowledge, there exists no gradient boosting algorithm that is applied to survival trees. On the other hand, studies on the Chinese consumer finance market \citep{li2019heterogeneous} indicate that ensemble methods outperform the traditional linear methods and the deep neural network as well, due to the characteristics of the credit data collected in Chinese market. This inspires us to develop a gradient boosting algorithm for survival tree models that optimizes the negative log-likelihood function. Hence, we propose in this paper a nonparametric ensemble tree model called \emph{gradient boosting survival tree} (GBST) that extends the survival tree models with a gradient boosting algorithm. The proposed model optimizes the survival probability of each time period simultaneously, which can reduce the overall error of the model significantly. 

The novelty of this paper is twofold. First, in order to deal with high-dimensional and heterogeneous credit data, we propose a gradient boosting algorithm to survival trees with likelihood as the optimization objective. Sepcifically, most of the boosting algorithms (e.g., \citealp{binder2008allowing, chen2013gradient}) applied to survival analysis are based on linear Cox model rather than trees, and meanwhile the tree-based survival models (\citealp{ishwaran2008random, wright2017unbiased}) employ the empirical log-rank split for generating trees rather than the split rule of optimizing likelihood objective. Second, we conduct experiments on two real large-scale credit datasets, where a comprehensive comparison is made between GBST and other popular time-to-event models, which fills the gap that most of literature in survival analysis methods are tested on small-scale clinical data.



The remainder of this paper is organized as follows. Section 2 introduces the methodology of gradient boosting survival tree (GBST). We start with a brief introduction of survival analysis, followed by a calculation of the negative log-likelihood loss function based on the hazard function represented by a survival tree ensemble. The loss function is optimized by a gradient boosting approach and the corresponding algorithms are derived. In Section 3, empirical experiments are conducted on two datasets: one is the lending club loan dataset that can be publicly retrieved from Kaggle and the other is a private dataset collected in Chinese consumer finance market. The results show that GBST outperforms the other candidate models (Cox, RSF, CoxBoost, XGB, CIF, GBMCI and DeepHit) on both datasets measured by the concordance index (C-index), Kolmogorov-Smirnov (KS) index, as well as by the area under the receiver operating characteristic curve (AUC) of each time period. Finally, we summarize this paper in Section 4.

\section{Gradient boosting survival tree}

In this section, we introduce the method of \emph{gradient boosting survival tree} (GBST) that extends the traditional survival analysis models with a gradient boosting tree algorithm.

\subsection{Survival analysis}
\label{subsection:sa}

The survival analysis aims, in general, for quantifying the distribution of time until some random event happens. In the field of consumer finance, the event is that a borrower defaults. Let $T$ denote the default time, i.e., the time period from the moment that an borrower obtains the money to the event that he defaults. Note that in general $T$ is a random variable and we assume it follows a probability distribution $\mathbb P$. Then for any time $t$, the survival function $S(t)$ is defined as follows,
\begin{equation*}
S(t) := \mathbb P(T>t),
\end{equation*}
which quantifies the probability that the borrower ``survives'' longer than time $t$. Comparing with the traditional credit scoring methods, the survival analysis is able to predict not only the probability whether a borrower defaults, but also the probability of time when he defaults. 

In the field of consumer finance, we need to consider the right censoring cases, i.e., the borrower does not default within the whole observation period, which is said to be ``censored''. An indicator $\delta$ is used to denote whether the sample is censored ($\delta = 0$) or not ($\delta = 1$). 

Throughout this paper, we consider a finite set of observation time, denoted by $\{\tau_j, j=0, 1, 2, \ldots, J\}$ satisfying $0 = \tau_0 < \tau_1 < \ldots < \tau_J < \infty$. In the context of consumer finance, the observation time could be, e.g., the monthly repayment due date. The discrete-time hazard function is defined as
\begin{align*}
 h(\tau_j) := \mathbb P( \tau_{j-1} < T \leq \tau_j | T > \tau_{j-1}), \quad j = 1, 2, \ldots,
\end{align*}
which quantifies the probability of the event that occurs within the time period $( \tau_{j-1}, \tau_j]$ for the first time. By definition, the survival function $S$ can be represented by the hazard function $h$. Indeed, for each $j = 1, 2, \ldots$, we have
\begin{align}
 S(\tau_j) = & \mathbb P(T>\tau_j) = \mathbb P(T>\tau_j | T > \tau_{j-1}) \mathbb P(T>\tau_{j-1} | T > \tau_{j-2}) \cdots \mathbb P(T > \tau_1) \nonumber \\
        = & (1 - h(\tau_j)) (1-h(\tau_{j-1})) \cdots (1-h(\tau_1))= \prod_{l=1}^j (1 - h(\tau_l)). \label{eq:hazard}
\end{align}
Finally, the probability of the event that occurs within the time period $( \tau_{j-1}, \tau_j]$ is given by
\begin{align}
 \mathbb P(\tau_{j-1} < T \leq \tau_j) = h(\tau_j) S(\tau_{j-1}) = h(\tau_j)\prod_{l=1}^{j-1} \left(1 - h(\tau_l) \right). \label{eq:likelihood}
\end{align}

Suppose that $h(t) \in (0, 1)$ for each $t$ and we consider a logarithmic transformation of $h$ such as
\begin{align}
 f(t) := & \log \frac{h(t)}{1 - h(t)},   \textrm{ satisfying }\\
  h(t) = & \frac{1}{1 + e^{-f(t)}} \textrm{ and } 1 - h(t) = \frac{1}{1 + e^{f(t)}}. \label{eq:h}
\end{align}
By defining the following two mappings according to the censoring indicator:
\begin{align}
 y_j(t) := \left\{ \begin{array}{ll}
                   -1, & \textrm{ if } \delta = 0 \ or \ (\delta = 1 \ and \ t > \tau_j), \\
                   1, & \textrm{ if } \delta = 1 \ and \ t \leq \tau_j,
                  \end{array}
 \right., \textrm{ and } \label{eq:y}
\end{align}
\begin{align*}
 J(t) := \left\{ \begin{array}{ll}
                 j, & \textrm{if } t \in (\tau_{j-1}, \tau_j], \\
                 J + 1, & \textrm{if } t > \tau_J.
                \end{array}
 \right.
\end{align*}
Eq.\ \eqref{eq:likelihood} implies
\begin{align*}
 \mathbb P( T=t) = \prod_{j=1}^{J(t)} \frac{1}{1 + e^{-y_j(t)f(\tau_j)}}, \forall t \leq \tau_J.
\end{align*}
On the other hand, if $t > \tau_J$, i.e., beyond the total observation period, its probability could be quantified as
\begin{align*}
 \mathbb P(T = t > \tau_J) =  \prod_{j=1}^{J} \frac{1}{1 + e^{f(\tau_j)}} = \prod_{j=1}^{J} \frac{1}{1 + e^{-y_j(t)f(\tau_j)}}.
\end{align*}
Combining the above two cases, we have that for any observed default time $t \in [0, \infty)$, its probability is given by
\begin{align}
 \mathbb P( T=t) = \prod_{j=1}^{J(t) \wedge J} \frac{1}{1 + e^{-y_j(t)f(\tau_j)}},\label{eq:likelihood2}
\end{align}
where $J(t) \wedge J := \min(J(t), J)$.

\noindent{\it Remark.} In many real applications, the exact ``death'' time might not be observable. Instead, with a discrete series of observation (sampling) time, one can usually determine in which period $(\tau_{j-1}, \tau_j]$ one individual is ``dead''. Then the output of $y_j(t)$ and $J(t)$ are observable without knowing the exact ``death'' time.

\subsection{Learning objective}

Let $\boldsymbol x \in \mathbb R^n$ denote the feature of each individual. In the field of consumer finance, it quantifies the characteristics of each borrower, such as age, sex, education level, etc. The fundamental task of survival analysis is then to investigate the optimal survival function for each individual $S(t; \boldsymbol x)$. As shown in Eq.\ \eqref{eq:hazard}, it is sufficient to find out the best hazard function $h(t; \boldsymbol x)$ and equivalently its logarithmic representation $f(t; \boldsymbol x)$.

Now we state the major assumption of this paper: $f$ is assumed to be approximated by $\hat f: [0, \infty) \times \mathbb R^n \rightarrow \mathbb R$ with a tree ensemble model such as
\begin{align}
 f(t; \boldsymbol x) \cong \hat f (t; \boldsymbol  x) := \sum_{k=1}^K f_k(t; \boldsymbol x), f_k \in \mathcal F,
\end{align}
where for each $k$, $f_k$ is the output of $k$th \emph{survival tree}, and
$$\mathcal F := \left\{ f(t; \boldsymbol x) = w(t, q(\boldsymbol x)) |  q: \mathbb R^n \rightarrow L , w: [0, \infty) \times L \rightarrow \mathbb R \right\}$$
is the space of survival trees with leaf nodes as functionals of time $t$. Here $q$ represents the structure of each tree with $L$ leaf nodes that map a variable $\boldsymbol x$ to its corresponding leaf index.  Each $f_k$ represents an independent tree structure $q_k$ and a leaf weight functional $w^{(k)}$. The difference between a survival tree and a conventional regression tree is that the leaf nodes of a survival tree are functionals of time $t$, whereas the leaf nodes of a regression tree are simply real values.


Suppose we observe samples $\{\boldsymbol x_i, t_i \}, i=1, 2, \ldots, N$, where $t_i$ denotes the default time. Under the i.i.d.\ assumption, we can calculate the negative log-likelihood as in Eq.\ \eqref{eq:likelihood2} as follows,
\begin{align}
\mathcal L(\phi) = & -\log \left( \prod_{i=1}^N \mathbb P(t_i; \boldsymbol x_i) \right) = - \log \left( \prod_{i=1}^N \prod_{j=1}^{J(t_i) \wedge J} \frac{1}{1 + e^{-y_j(t_i) f (\tau_j; \boldsymbol x_i)}} \right) \nonumber \\
 \cong & - \log \left( \prod_{i=1}^N \prod_{j=1}^{J(t_i) \wedge J} \frac{1}{1 + e^{-y_j(t_i) \hat f (\tau_j; \boldsymbol x_i)}} \right) \nonumber \\
 = & \sum_{i=1}^N \sum_{j=1}^{J(t_i) \wedge J} \log \left( 1 + \exp \left\{ - y_j(t_i) \hat f(\tau_j; \boldsymbol x_i) \right\} \right). \label{eq:l}
\end{align}
For each $j=1, 2, \ldots, J$, the set of samples that survive longer than $\tau_{j-1}$ (in other words, $J(t_i) \geq j$), $N_j$, is defined as follows:
\begin{align}
 N_j := \{ i \in \{1, 2, \ldots, N\} | J(t_i) \geq j \}. \label{eq:nj}
\end{align}
Then it is easy to verify that Eq.\ \eqref{eq:l} implies
\begin{align*}
 \mathcal L(\phi) = & \sum_{j=1}^J \sum_{i \in N_j}  \log \left( 1 + \exp\left(- y_j(t_i) \hat f(\tau_j; \boldsymbol x_i) \right) \right).
\end{align*}
Finally, in order to avoid over-fitting, one $l^2$ regularization term is added to penalize the complexity of the model:
\begin{align}
 \mathcal L(\phi) = &\sum_{j=1}^J \sum_{i \in N_j}  \log \left( 1 + \exp\left(- y_j(t_i) \hat f(\tau_j; \boldsymbol x_i) \right) \right) + \frac{\lambda}{2} \sum_k \lVert w^{(k)} \rVert^2, \label{eq:l2}
\end{align}
where $\lambda > 0$ is a hyperparameter controls how large the penalization is and $w^{(k)}$ is the leaf weight functionals for $k$th survival tree. It should be noted that since we consider in this paper merely the discrete observation time $\{ \tau_j \}$, the $l^2$-norm of a leaf weight functional $w$ is calculated as $\lVert w \rVert^2 := \sum_{j=1}^J \sum_{l=1}^L w_l^2(\tau_j).$


\subsection{Gradient boosting}

Inspired by the gradient boosting method \citep{friedman2001greedy}, the optimization problem stated in Eq.\ \eqref{eq:l2} is solved in an additive manner. More specifically, one starts with an initial \emph{survival tree} $f_0$ and for $m=1, 2, \ldots, $
\begin{align}
 f_m = \argmin_{f \in \mathcal F} \mathcal L^{(m)} := & \sum_{j=1}^J \sum_{i \in N_j} \log \left( 1 + \exp\left\{- y_j(t_i) \left( \hat f^{(m-1)}(\tau_j; \boldsymbol x_i) + f(\tau_j; \boldsymbol x_i) \right) \right\} \right) \nonumber \\
  & + \frac{\lambda}{2} \lVert w \rVert^2 \label{eq:gb}
\end{align}
and $\hat f$ is updated by
\begin{align*}
 \hat f^{(m)}(t; \boldsymbol x ) =  \hat f^{(m-1)}(t; \boldsymbol x ) + f_m(t; \boldsymbol x).
\end{align*}

Define the logarithmic loss function $\varrho : \mathbb R \times \mathbb R \rightarrow \mathbb R$ as
$$\varrho(y, f) := \log ( 1+ \exp(-y \cdot f)).$$
Then $\mathcal L^{(m)}$ can be approximated by a Taylor expansion up to the second order:
\begin{align}
 \mathcal L^{(m)}(f) \cong & \sum_{j=1}^J \sum_{i \in N_j} \left( \underbrace{\varrho\left(y_j(t_i), \hat f^{(m-1)}(\tau_j; \boldsymbol x_i)\right)}_{\textrm{independent of } f} + r^{(m-1)}_{i,j} f(\tau_j; \boldsymbol x_i) + \frac{1}{2} \sigma^{(m-1)}_{i,j} f^2(\tau_j; \boldsymbol x_i) \right) \nonumber\\
  & + \frac{\lambda}{2} \lVert w \rVert^2, \label{eq:lm}
\end{align}
where the first and second order derivative $r^{(m-1)}_{i,j}$ and $\sigma^{(m-1)}_{i,j}$ are given by
\begin{align*}
r^{(m-1)}_{i,j} := \frac{\partial \varrho}{\partial f} \left(y_j(t_i), \hat f^{(m-1)}(\tau_j; \boldsymbol x_i)\right), \\
\sigma^{(m-1)}_{i,j} := \frac{\partial^2 \varrho}{\partial f^2} \left(y_j(t_i), \hat f^{(m-1)}(\tau_j; \boldsymbol x_i)\right).
\end{align*}
Removing the constant independent of $f$ in Eq.\ \eqref{eq:lm}, the objective becomes
\begin{align}
 \tilde{\mathcal L}^{(m)}(f) := \sum_{j=1}^J \sum_{i \in N_j} \left( r^{(m-1)}_{i,j} f(\tau_j; \boldsymbol x_i) + \frac{1}{2} \sigma^{(m-1)}_{i,j} f^2(\tau_j; \boldsymbol x_i) \right) + \frac{\lambda}{2} \lVert w \rVert^2, \label{eq:tilde}
\end{align}

Note that since $y_j(t_i)$ takes value either 1 or -1 (see Eq.\ \eqref{eq:y}), for the logarithmic loss function $\varrho$, it is easy to verify that
\begin{align}
 \frac{\partial \varrho}{\partial f}(y, f) = & \left\{
\begin{array}{ll}
-(1 - h), & \textrm{ if } y = 1\\
h, & \textrm{ if } y = -1
\end{array}
 \right., \textrm{ and }\\
 \frac{\partial^2 \varrho}{\partial f^2}(y, f) = & h(1 - h),  \textrm{ if }  y = 1 \textrm{ or} -1, \label{eq:h2}
\end{align}
where $h :=  \frac{1}{1 + e^{-f}}$ can be interpreted as the hazard probability (cf.\ Eq.\ \eqref{eq:h}).

For a survival tree with $L$ leaf nodes, $f$ can be represented as
$$f(\tau_j; \boldsymbol x_i) = \sum_{l=1}^L w_l(\tau_j) \mathbf 1 (i \in I_l)$$
where we define $I_l := \{ i \in \{1, 2, \ldots, N \} | q(\boldsymbol x_i) = l\}$ and $q$ represents the structure of the tree. Eq.\ \eqref{eq:tilde} becomes
\begin{align*}
 \tilde{\mathcal L}^{(m)}(w) := \sum_{j=1}^J \sum_{l=1}^L \left\{ \sum_{i \in N_j \cap I_l} \left( r^{(m-1)}_{i,j}  w_l(\tau_j) + \frac{1}{2} \sigma^{(m-1)}_{i,j} w_l^2(\tau_j)  \right) + \frac{\lambda}{2} w_l^2(\tau_j) \right\}.
\end{align*}
As shown in Eq.\ \eqref{eq:h2}, $\sigma^{(m-1)}_{i,j} > 0, \forall i, j,$ and $\tilde{\mathcal L}^{(m)}$ is therefore a strictly convex function with respect to $w$. Hence, given a fixed structure $q$, for each $j$ and $l$, the optimal $ w_l^{(m)}(\tau_j)$ is
\begin{align}
 w_l^{(m)}(\tau_j) = - \frac{\sum_{i \in N_j \cap I_l} r^{(m-1)}_{i, j}}{\sum_{i \in N_j \cap I_l} \sigma^{(m-1)}_{i, j} + \lambda}, \label{eq:w}
\end{align}
and the corresponding $\tilde{\mathcal L}^{(m)}$ is
\begin{align}
 \tilde{\mathcal L}^{(m)} = & - \frac{1}{2} \sum_{j,l} \frac{\left( \sum_{i \in N_j \cap I_l} r^{(m-1)}_{i, j} \right)^2}{\sum_{i \in N_j \cap I_l} \sigma^{(m-1)}_{i, j} + \lambda}, \label{eq:score}
\end{align}
where the explicit form of $r^{(m-1)}_{i, j} $ and $\sigma^{(m-1)}_{i, j}$ is
\begin{align*}
 r^{(m-1)}_{i, j}  = & \left\{ \begin{array}{ll}
                                -(1-h_{i, j}^{(m-1)}), & \textrm{if } y_j(t_i)=1\\
                                h_{i, j}^{(m-1)}, & \textrm{if } y_j(t_i)=-1
                               \end{array}
 \right. \\
 \sigma^{(m-1)}_{i, j}  =& (1-h_{i, j}^{(m-1)}) h_{i, j}^{(m-1)} ,
\end{align*}
with
\begin{align*}
 h_{i, j}^{(m-1)} := \frac{1}{1 + \exp(- \hat f^{(m-1)}(\tau_j; \boldsymbol x_i))}
\end{align*}
being the hazard probability estimated for $\boldsymbol x_i$ at time $\tau_j$ up to $(m-1)$th iteration.

Eq.\ \eqref{eq:score} measures how well a survival tree $q$ fits the observed data. Based on it, we develop a greedy algorithm for constructing the optimal survival tree, since usually it is impossible to explore all possible tree structures. We start with a single leaf node and iteratively split it into two sets. Suppose that one set $I$ is split into two separate sets $I_L$ and $I_R$. Then the corresponding loss reduction is given by
\begin{align}
 \tilde{\mathcal L}_{split} = \frac{1}{2} \sum_{j}\left[ \frac{\left( \sum_{i \in N_j \cap I_L} r^{(m-1)}_{i, j} \right)^2}{\sum_{i \in N_j \cap I_L} \sigma^{(m-1)}_{i, j} + \lambda}+ \frac{\left( \sum_{i \in N_j \cap I_R} r^{(m-1)}_{i, j} \right)^2}{\sum_{i \in N_j \cap I_R} \sigma^{(m-1)}_{i, j} + \lambda}- \frac{\left( \sum_{i \in N_j \cap I} r^{(m-1)}_{i, j} \right)^2}{\sum_{i \in N_j \cap I} \sigma^{(m-1)}_{i, j} + \lambda}\right].
\end{align}
The pseudo code of the proposed greedy split algorithm is shown in Algorithm \ref{alg:split}. Some techniques to speed up the split algorithm is demonstrated in Appendix.

\begin{algorithm}[!h]
  \caption{Split algorithm for GBST.}
  \label{alg:split}
  \begin{algorithmic}[1]
    \Require
     $I$: Set of individuals in current node; $\left\{ \boldsymbol x_i \in \mathbb R^n \right\}$: the features of individuals; $N_j, j = 1, 2, \ldots, J$ defined in Eq.\ \eqref{eq:nj}
    \For{$k=1$ to $K$}
    \For{$j=1$ to $J$}
    \State $W_j\leftarrow \sum_{i \in N_j \cap I} r_{i, j}$, $V_j\leftarrow \sum_{i \in N_j \cap I} \sigma_{i, j}$
    \EndFor
    \For{$s$ in sorted($I$, by $\boldsymbol x_{sk}$)}
    \For{$j=1$ to $J$}
    \If{$s \in N_j$}
    \State $W_{jL}\leftarrow W_{jL}+r_{s, j}$, $V_{jL}\leftarrow V_{jL}+\sigma_{s, j}$
    \State $W_{jR}=W_j-W_{jL}$, $V_{jR}=V_j-V_{jL}$
    \EndIf
    \EndFor
    \State $score \leftarrow \max(score, \sum_j[\frac{W_{jL}^2}{V_{jL}+\lambda}+\frac{W_{jR}^2}{V_{jR}+\lambda}-\frac{W_{j}^2}{V_{j}+\lambda}]$)
    \EndFor
    \EndFor
    \Ensure
      Split with max score, splitting schema
  \end{algorithmic}
\end{algorithm}

To initialize the tree, one has to give an initial estimation of the hazard probability $h_{i, j}^{(0)}$. Without any prior knowledge, we assume each individual follows the same hazard  distribution over time with the Kaplan-Meier estimator \citep{kaplan1958nonparametric}. More specifically, let $d_j$ be the number of default individuals at $(\tau_{j-1}, \tau_j]$ and $n_j$ be the number of individuals whose survival time is larger than $\tau_{j-1}$, i.e., formally,
\begin{align*}
 d_j := \sum_{i=1}^N \mathbf 1(J(t_i) = j),  \quad n_j := \sum_{i=1}^N \mathbf 1(J(t_i) \geq j),
\end{align*}
where $t_i$ is the default time of $i$th individual. Then, the hazard probability by the Kaplan-Meier estimator is calculated as
\begin{align}
 \hat h_j = \frac{d_j}{n_j}, \label{eq:h0}
\end{align}
which is used as our initial estimation $h_{i, j}^{(0)} = \hat h_j, \forall i, j. $

Finally, we summarize the algorithm for GBST in Algorithm \ref{alg:framework}. All the codes of GBST are publicly available in Github\footnote{\url{https://github.com/360jinrong/GBST}}.
\begin{algorithm}[!h]
  \caption{Algorithm for GBST}
  \label{alg:framework}
  \begin{algorithmic}[1]
    \Require
    $\boldsymbol x \in \mathbb R^n$: the features of individuals; $y_j(t_i), j=1,...,J; i=1,...,N$: the default labels of individuals at observed times; $M$: the max.\ number of trees or iterations.
    \State Calculate the set $N_j$ for each $j$ defined in Eq.\ \eqref{eq:nj}.
    \State Initialize $h_{i, j}^{(0)} = \hat h_j, \forall i, j$ as in Eq.\ \eqref{eq:h0}.
    \For {$m=1$ to $M$}
    \State Construct a tree with the split algorithm (see Algorithm \ref{alg:split})
    \State In each terminal node $l$, calculate $w_l^{(m)}$ as in Eq.\ \eqref{eq:w} for each $l$
    \State Calculate $f_m(\tau_j;\boldsymbol x_i) \leftarrow \sum_{l} w_l(\tau_j) \mathbf 1 (i \in I_l)$
    \State Update $\hat f^{(m)}(\tau_j;\boldsymbol x_i)\leftarrow \hat f^{(m-1)}(\tau_j;\boldsymbol x_i)+f_m(\tau_j;\boldsymbol x_i)$
    \EndFor
    \State Calculate the hazard function $h(\tau_j;\boldsymbol x_i)\leftarrow \frac{1}{1+ \exp \left\{-\hat f^{(M)}(\tau_j;\boldsymbol x_i)\right\}}$
    \Ensure
      The Survival function $S(\tau_j;\boldsymbol x_i)\leftarrow \prod_{l=1}^{j} \left(1 - h(\tau_l;\boldsymbol x_i)\right )$
  \end{algorithmic}
\end{algorithm}

\section{Experiments}

In order to show the applicability of the proposed gradient boosting survival tree (GBST) algorithm, we apply it to analyze two datasets. One is the \emph{lending club loan data}\footnote{\url{https://www.kaggle.com/sonujha090/xyzcorp-lendingdata?select=XYZCorp_LendingData.txt}}, which is publicly available from Kaggle, and the other is a private loan dataset provided by 360 Finance Inc. It is noteworthy that the former public dataset is collected from the US market, since to our best knowledge, there exists no public dataset available from Chinese consumer finance market. 

We compare the performance of GBST with the performance of other existing models: Cox proportional hazards model (COX), random survival forest (RSF, \citealp{ishwaran2008random}), XGBoost (XGB, \citealp{chen2016xgboost}), CoxBoost \citep{binder2008allowing}, gradient boosting machine for concordance index (GBMCI, \citealp{chen2013gradient}), conditional inference forest model (CIF, \citealp{wright2017unbiased}) and DeepHit \citep{lee2018deephit}. 

Credit scoring is usually solved by classification methods. It is, therefore, of great interest to see comparison of the proposed GBST model with classification methods. Among them, XGB dominates the Kaggle competitions on tabular data \citep{kaggle2019}. Therefore, we choose XGB as the best candidate among all classification methods.

There are several open-source packages implemented for the models mentioned above. Among them, there are implementations in R for CoxBoost, GBMCI and ICF, while Python packages exist for the other models. For readers' convenience, the link addresses of the implemented packages of the models are listed in Appendix \ref{app:links}. We also notice that lately there exists an in-progress Python package \citep{sksurv} called ``gradient boosting survival analysis'', which provides several implementations of popular models, including Cox, RSF and CoxBoost, as well as a Python package ``PyCox'' by \cite{geck2012pycox}, which provides in addition an implementation of DeepHit.


\subsection{Lending club loan data}
\label{sec:lcld}

\subsubsection{Dataset} 

The lending club loan dataset retrieved from Kaggle contains data of all loans issued from 2007 to 2015, including the loan status (normal, late, fully paid) and payment information. It totally contains 855,527 individual loans. The observation period we selected is 24 months. 124,219 loans issued before May 2013 are selected as the training set and 100,188 loans issued between May 2013 and December 2013 as the test set (see Table \ref{table:lcldata}). Since the data ends at end of 2015, The loans issued after 2013 are not used due to the incomplete observation period which is less than 24 months. 

\begin{table}[!ht]
	\centering
	\begin{tabular}{c|c|c}
		\hline
		{\bf dataset}& \bf issued month& \bf sample size\\
		\hline
		training set& January 2007 to April 2013 & 124,219\\
		test set& May 2013 to December 2013 & 100,188\\
		\hline
	\end{tabular}
	\caption{Summary of datasets.}
	\label{table:lcldata}
\end{table}

\begin{figure}[!ht]
  \centering
    \includegraphics[width=1\textwidth]{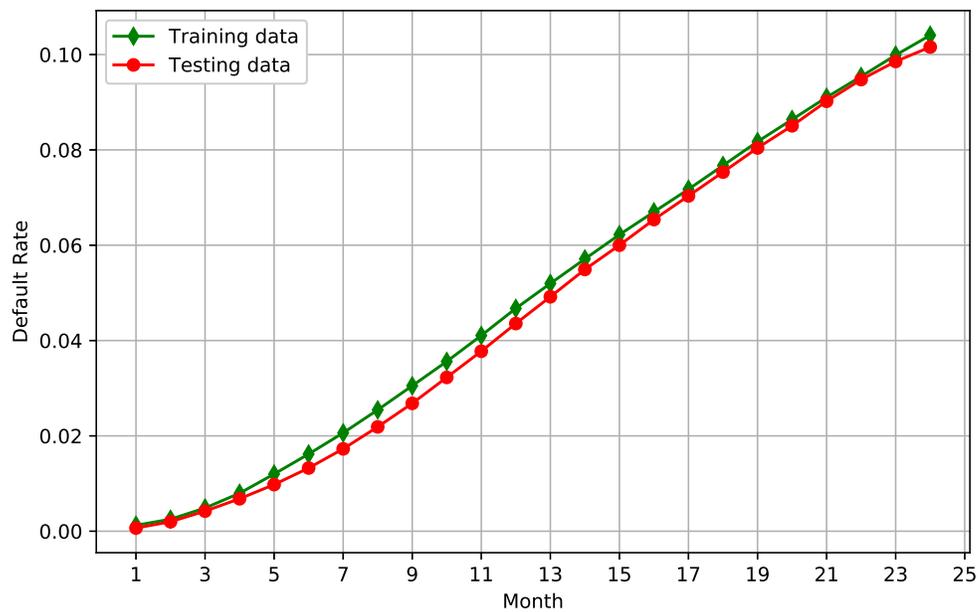}
   \caption{Lending club loan data: (cumulative) default rates (Eq.\ \eqref{eq:dr}) for both the training and test dataset.}
   \label{fig:defaultlc}
\end{figure}

 Based on the last month payment (labeled by ``last\_pymnt\_d'' in the dataset), issued month (``issue\_d'') and the default indicator\footnote{The default indicator is labeled by ``default\_ind'' and is taken value 0 or 1, whereas the definition of default is not specified in the description of the dataset.}, we can calculate for each loan the survival time (or censoring time) and the censoring indicator. In this paper, we assume that if a loan defaults in some repayment month, it remains in the default state in the rest months, regardless of later repayments. The \emph{default rate} in certain repayment month $t$ is defined to be the number of loans that default in this month or were so in previous months over the total number of accounts, i.e.,
\begin{align}
 \textrm{default rate}(t) = \frac{\# \textrm{default accounts up to month } t }{\# \textrm{total accounts}}. \label{eq:dr}
\end{align}

Note that the default rate defined here is accumulative over time, in accordance with default probability used in survival analysis. Hence, the curve of default rate grows with time, as depicted in Figure \ref{fig:defaultlc}. We can also see in Figure \ref{fig:defaultlc} that the default rate curve of the training data is almost identical with that of the test data.

\subsubsection{Pre-processing of data} 
\label{sec:preprocessing}
In the lending club loan dataset, there are 55 variables for each loan, including its ID and default label. Among these variables, the ID-like variables and the variables with a missing rate higher than 80\% are first excluded. Categorical variables, e.g., the loan purpose and employment length, are represented by numerical values with one-hot encoding. For each variable, the missing value is filled up with a default value and an additional binary variable is generated to indicate whether the value is missing. Eventually 42 variables among 55 are chosen as features for model inputs. The selected features mainly include:
\begin{itemize}
  \item loan information, e.g., the loan amount, interest rate, loan grade;
  \item personal information, e.g., home ownership, the number of finance trades;
  \item historical credit information, e.g., the total number of credit lines in the borrower's credit file, etc.
\end{itemize}

\subsubsection{Convergence of GBST}

In this subsection, we demonstrate that GBST algorithm converges after dozens of iterations when we apply the algorithm to the lending club loan data. Note that at each iteration of the GBST algorithm (see Algorithm \ref{alg:framework}), a survival tree is constructed by minimizing the loss function stated in Eq.\ \eqref{eq:l2} on the training dataset. This turns out to be very time-costly when the dataset is large. Hence, we apply a sampling approach as proposed in stochastic gradient boosting \citep{friedman2002stochastic} to reduce the computational cost. Instead of using the entire training dataset, we randomly sample a subset of the training dataset and apply them to train the model, with certain sampling rate $\alpha$. Note that the loss defined in Eq.\ \eqref{eq:l2} measures how well the proposed GBST model fits the observed samples. To test the convergence behavior of the GBST algorithm with random sampling, we run the algorithm 1000 times independently. In this experiment, we use the sampling rate $\alpha= 20\%$ and the regularization parameter $\lambda = 0.001$, and the max tree depth is set to $6$. 

\begin{figure}[!ht]
  \centering
    \includegraphics[width=0.8\textwidth]{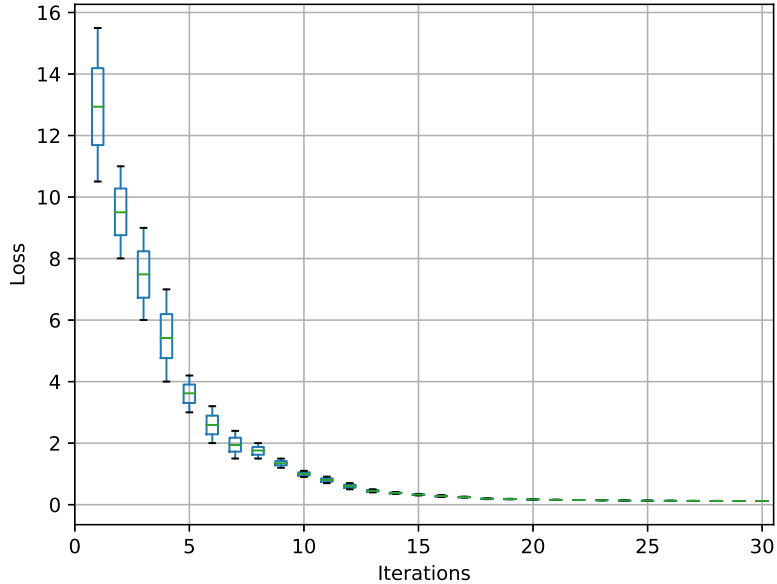}
   \caption{Distribution of loss values defined in Eq.\ \eqref{eq:l2} over iterations with 1000 independent runs on the lending club loan data.}
   \label{fig:loss}
\end{figure}

Figure \ref{fig:loss} plots the distribution of loss values over iterations of these 1000 independent runs. For each iteration, we calculate the minimal, maximal, 25\% quantile, median and 75\% quantile of the 1000 independent loss values and plot in Figure \ref{fig:loss} with a box plot. Figure \ref{fig:loss} shows that the loss values decrease very quickly over iterations and converges uniformly to a certain level near 0, as the iteration grows. Hence, the GBST algorithm with random sampling shows a promising property of convergence. It shows also that 30 iterations are sufficient to obtain the optimal solution. 

\subsubsection{Performance of GBST}
\label{sec:performance}
\begin{figure}[!ht]
  \centering
    \includegraphics[width=0.8\textwidth]{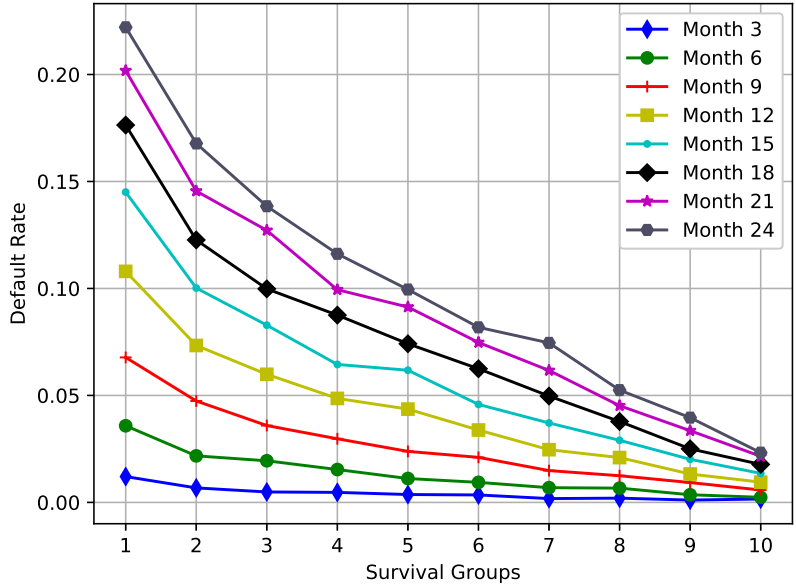}
   \caption{Default rates defined in Eq.\ \eqref{eq:dr} of different survival groups in the test dataset for every 3 repayment months. Samples in the test dataset are sorted in ascending order according to the survival probabilities of each month predicted by the GBST, i.e., the larger the group index is, the higher the predicted survival probabilities of the group are.}
   \label{fig:default2lc}
\end{figure}

We apply the GBST model to calculate the predicted survival probability of each loan in the test dataset for every repayment month. The loans are sorted in ascending order according to the predicted survival probabilities and are divided into 10 survival groups with equal sample size. Hence, the larger the group index is, the higher the predicted survival probabilities within the group are. For each survival group, we calculate the true default rate defined in Eq.\ \eqref{eq:dr} and the results are plotted in Figure \ref{fig:default2lc}. It shows that for each repayment month (labeled by different shapes and colors in Figure \ref{fig:default2lc}) the true default rate decreases monotonically as the predicted survival probability increases, which indicates that the GBST model has a good discriminating power in the test dataset over all months.

To examine the model performance more rigorously, we adopt the concordance index (C-index) proposed by \cite{harrell1982evaluating} to measure the goodness of fit for survival models. C-index provides a global assessment of the model for continuous events. It takes values ranging from 0.5 (no discrimination) to 1.0 (perfect discrimination). The value of C-index is 0.6740 for GBST, showing that the GBST model has a certain ability to distinguish risky loans in the lending club loan dataset.

\subsubsection{Comparison}

Other existing models (COX, XGB, RSF, CoxBoost, GBMCI, CIF and DeepHit) are also employed to analyze the lending club loan dataset, and we compare the results of GBST with the results of the other models under several performance metrics. For all models, we apply the same procedure of data pre-processing as described in Section \ref{sec:preprocessing}. It should be noted that XGB is a classification model and it cannot predict survival probability curves. Nevertheless, we use the following label as the learning target of XGB model: a loan is labeled with ``default'' if it defaults in any repayment month within the observation period of 24 months, and with ``non-default'' otherwise. After training, XGB model is able to predict non-default probability for each loan, which can then be used to calculate the performance metrics to be introduced below. 

In order to make the comparison statistically solid, we conduct 10 leave-one-out cross-validation (LOOCV) experiments: all data is divided into 10 almost equal-size datasets according to the issued months (see Table \ref{table:lcldata2}), and for each of the 10 experiments, 9 of them are combined as the training set while the remaining one as the test set. In each experiment, the hyper\-parameters of all candidate models are optimized in the training dataset by a random grid search with 5-fold cross validation, which is used to avoid over-fitting. For instance, the hyperparameters of the GBST model to be optimized are the maximum tree depth, learning rate, number of trees and sub-sample ratio. The optimization objective for XGB is AUC (the area under the receiver operating characteristics curve), while for the other survival models including GBST the objective is C-index. In other words, we ensure that each candidate model is optimized separately and a fair comparison is guaranteed. 

 
 \begin{table}[!ht]
	\centering
	\begin{tabular}{c|c|c}
		\hline
		{\bf index}& \bf issued month& \bf sample size\\
		\hline
		1 & June 2007 to March 2011 & 23,726\\
		2 & April 2011 to February 2012 & 24,013\\
		3 & March 2012 to July 2012 & 20,399\\
		4 & August 2012 to November 2012 & 24,591\\
		5 & December 2012 to February 2013 & 22,278\\
		6 & March 2013 to May 2013 & 29,996\\
		7 & June 2013 to July 2013 & 24,046\\
		8 & August 2013 & 12,694\\
		9 & September 2013 to October 2013 & 28,055\\
		10 & November 2013 to December 2013 & 14,609\\
		\hline
	\end{tabular}
	\caption{10 datasets of Lending Club data by issued months.}
	\label{table:lcldata2}
\end{table}

\setlength{\tabcolsep}{2mm}{
\begin{table}[]
	\centering
	\begin{threeparttable}
	\begin{tabular}{|c|c|c|c|c|}
		\hline
		Algorithm & GBST                                                                           & COX                                                                   & RSF                                                                   & XGB                                                                   \\ \hline
		C-index   & \textbf{\begin{tabular}[c]{@{}c@{}}0.6867\\      (0.6735-0.6999)\end{tabular}} & \begin{tabular}[c]{@{}c@{}}$0.6799^+$\\      (0.6624-0.6934)\end{tabular} & \begin{tabular}[c]{@{}c@{}}$0.6666^*$\\      (0.6455-0.6877)\end{tabular} & \begin{tabular}[c]{@{}c@{}}$0.6809^+$\\      (0.6665-0.6953)\end{tabular} \\ \hline
		Algorithm & CoxBoost                                                                       & GBMCI                                                                 & CIF                                                                   & DeepHit                                                               \\ \hline
		C-index   & \begin{tabular}[c]{@{}c@{}}$0.6779^{++}$\\      (0.6670-0.6888)\end{tabular}          & \begin{tabular}[c]{@{}c@{}}$0.6742^*$\\      (0.6561-0.6923)\end{tabular} & \begin{tabular}[c]{@{}c@{}}$0.6694^*$\\      (0.6455-0.6933)\end{tabular} & \begin{tabular}[c]{@{}c@{}}$0.6745^*$\\      (0.6587-0.6903)\end{tabular} \\ \hline
	\end{tabular}
	\begin{tablenotes}
		\footnotesize
		\item[$*$] indicates p-value $ < 0.001$
		\item[$+$] indicates p-value $ < 0.01$
		\item[$++$] indicates p-value $ < 0.05$
	\end{tablenotes}
	\end{threeparttable}
	\caption{C-indices of 10 leave-one-out cross validatiton experiments (in terms of mean values and 95\% confidence intervals).}
	\label{table:cindex_n2}
\end{table}
}

First, we calculate the C-indices for all models and Table \ref{table:cindex_n2} summarizes the C-indices in the 10 LOOCV experiments (in terms of mean values and 95\% confidence intervals), which shows that all candidate models show some extent of discriminating power with C-index ranging from 0.6666 to 0.6867, while GBST achieves the highest C-index among all models. In order to test the significance of the differences, we conduct a T-test with the 10 C-indices on GBST and other methods. Table \ref{table:cindex_n2} shows that GBST outperforms other models significantly with all p-values $< 0.05$.

Second, the AUC and Kolmogorov-Smirnov (KS) index are two widely used performance metrics for binary classification methods in credit scoring \citep{thomas2002credit}. Note that most of the survival models are able to predict the survival probability of each loan for each repayment month, and therefore, we can calculate AUC and KS indices for each repayment month in the 10 LOOCV experiments (in terms of mean values and 95\% confidence intervals). For each repayment month, we only take those ``survived'' loans into account in the calculation of AUC and KS index, and hence, the samples size decreases as time increases. It is also noteworthy that since GBMCI can only predict estimated survival time rather than survival probability, KS index and AUC are not computable by GBMCI. 

\setlength{\tabcolsep}{7mm}{
\begin{table}[]
\centering
\begin{threeparttable}
\resizebox{\textwidth}{!}{%
\begin{tabular}{|c|c|c|c|c|c|c|c|c|}
	\hline
	Month               & Index & GBST                                                                       & XGB                                                               & COX                                                              & RSF                                                              & CoxBoost                                                                  & CIF                                                              & DeepHit                                                           \\ \hline
	\multirow{2}{*}{1}  & AUC   & \textbf{\begin{tabular}[c]{@{}c@{}}0.7702\\ (0.7544-0.7861)\end{tabular}}  & \begin{tabular}[c]{@{}c@{}}$0.7389^*$\\ (0.72169-0.7562)\end{tabular} & \begin{tabular}[c]{@{}c@{}}$0.7132^*$\\ (0.6946-0.7318)\end{tabular} & \begin{tabular}[c]{@{}c@{}}$0.6639^*$\\ (0.6386-0.6892)\end{tabular} & \begin{tabular}[c]{@{}c@{}}$0.7393^{++}$\\ (0.7262-0.7524)\end{tabular}          & \begin{tabular}[c]{@{}c@{}}$0.5972^*$\\ (0.5685-0.6259)\end{tabular} & \begin{tabular}[c]{@{}c@{}}$0.6923^*$\\ (0.6734-0.7113)\end{tabular}  \\ \cline{2-9} 
	& KS    & \textbf{\begin{tabular}[c]{@{}c@{}}0.4245\\ (0.4087-0.4404)\end{tabular}}  & \begin{tabular}[c]{@{}c@{}}$0.4010^*$\\ (0.3837-0.4183)\end{tabular}  & \begin{tabular}[c]{@{}c@{}}$0.3238^*$\\ (0.3052-0.3424)\end{tabular} & \begin{tabular}[c]{@{}c@{}}$0.2911^*$\\ (0.2658-0.3164)\end{tabular} & \begin{tabular}[c]{@{}c@{}}$0.3895^{++}$\\ (0.3764-0.4026)\end{tabular}          & \begin{tabular}[c]{@{}c@{}}$0.2594^*$\\ (0.2307-0.2881)\end{tabular} & \begin{tabular}[c]{@{}c@{}}$0.2586^*$\\ (0.2396-0.2775)\end{tabular}  \\ \hline
	\multirow{2}{*}{3}  & AUC   & \begin{tabular}[c]{@{}c@{}}0.7101\\ (0.6915-0.7288)\end{tabular}           & \begin{tabular}[c]{@{}c@{}}0.7159\\ (0.6955-0.7363)\end{tabular}  & \begin{tabular}[c]{@{}c@{}}$0.7004^+$\\ (0.6785-0.7224)\end{tabular} & \begin{tabular}[c]{@{}c@{}}$0.6631^*$\\ (0.6333-0.6930)\end{tabular} & \textbf{\begin{tabular}[c]{@{}c@{}}0.7185\\ (0.7030-0.7339)\end{tabular}} & \begin{tabular}[c]{@{}c@{}}$0.6726^+$\\ (0.6388-0.7065)\end{tabular} & \begin{tabular}[c]{@{}c@{}}$0.6875^{++}$\\ (0.6651-0.7099)\end{tabular}  \\ \cline{2-9} 
	& KS    & \begin{tabular}[c]{@{}c@{}}0.3002\\ (0.2819-0.3185)\end{tabular}           & \begin{tabular}[c]{@{}c@{}}0.3107\\ (0.2907-0.3306)\end{tabular}  & \begin{tabular}[c]{@{}c@{}}$0.3008^+$\\ (0.2793-0.3223)\end{tabular} & \begin{tabular}[c]{@{}c@{}}$0.2392^*$\\ (0.2099-0.2685)\end{tabular} & \textbf{\begin{tabular}[c]{@{}c@{}}0.3201\\ (0.3050-0.3352)\end{tabular}} & \begin{tabular}[c]{@{}c@{}}$0.3026^+$\\ (0.2694-0.3358)\end{tabular} & \begin{tabular}[c]{@{}c@{}}$0.2430^{++}$\\ (0.2210-0.2649)\end{tabular}  \\ \hline
	\multirow{2}{*}{5}  & AUC   & \textbf{\begin{tabular}[c]{@{}c@{}}0.7267\\ (0.7055-0.7480)\end{tabular}}  & \begin{tabular}[c]{@{}c@{}}$0.7252^*$\\ (0.7021-0.7484)\end{tabular}  & \begin{tabular}[c]{@{}c@{}}$0.7124^*$\\ (0.6875-0.7374)\end{tabular} & \begin{tabular}[c]{@{}c@{}}$0.6942^+$\\ (0.6602-0.7282)\end{tabular} & \begin{tabular}[c]{@{}c@{}}$0.7260^*$\\ (0.7084-0.7435)\end{tabular}          & \begin{tabular}[c]{@{}c@{}}$0.7131^*$\\ (0.6746-0.7516)\end{tabular} & \begin{tabular}[c]{@{}c@{}}$0.6862^{++}$\\ (0.6608-0.7116)\end{tabular}  \\ \cline{2-9} 
	& KS    & \textbf{\begin{tabular}[c]{@{}c@{}}0.3376\\ (0.3176-0.3576)\end{tabular}}  & \begin{tabular}[c]{@{}c@{}}$0.3343^*$\\ (0.3125-0.35612)\end{tabular} & \begin{tabular}[c]{@{}c@{}}$0.3075^*$\\ (0.2841-0.3310)\end{tabular} & \begin{tabular}[c]{@{}c@{}}$0.2834^+$\\ (0.2515-0.3153)\end{tabular} & \begin{tabular}[c]{@{}c@{}}$0.3320^*$\\ (0.3155-0.3485)\end{tabular}          & \begin{tabular}[c]{@{}c@{}}$0.3192^*$\\ (0.2830-0.3553)\end{tabular} & \begin{tabular}[c]{@{}c@{}}$0.2393^{++}$\\ (0.2154-0.2632)\end{tabular}  \\ \hline
	\multirow{2}{*}{7}  & AUC   & \textbf{\begin{tabular}[c]{@{}c@{}}0.7228\\ (0.6999-0.7457)\end{tabular}}  & \begin{tabular}[c]{@{}c@{}}$0.7207^+$\\ (0.6956-0.7457)\end{tabular}  & \begin{tabular}[c]{@{}c@{}}$0.7111^*$\\ (0.6842-0.7381)\end{tabular} & \begin{tabular}[c]{@{}c@{}}$0.6925^*$\\ (0.6557-0.7292)\end{tabular} & \begin{tabular}[c]{@{}c@{}}0.7206\\ (0.7017-0.7396)\end{tabular}          & \begin{tabular}[c]{@{}c@{}}$0.6938^*$\\ (0.6522-0.7353)\end{tabular} & \begin{tabular}[c]{@{}c@{}}$0.6806^*$\\ (0.6531-0.7081)\end{tabular}  \\ \cline{2-9} 
	& KS    & \textbf{\begin{tabular}[c]{@{}c@{}}0.3167\\ (0.2956-0.3378)\end{tabular}}  & \begin{tabular}[c]{@{}c@{}}$0.3162^+$\\ (0.2932-0.3393)\end{tabular}  & \begin{tabular}[c]{@{}c@{}}$0.3015^*$\\ (0.2767-0.3263)\end{tabular} & \begin{tabular}[c]{@{}c@{}}$0.2807^*$\\ (0.2469-0.3145)\end{tabular} & \begin{tabular}[c]{@{}c@{}}0.3137\\ (0.2962-0.3311)\end{tabular}          & \begin{tabular}[c]{@{}c@{}}$0.2987^*$\\ (0.2604-0.3370)\end{tabular} & \begin{tabular}[c]{@{}c@{}}$0.2308^*$\\ (0.2055-0.2561)\end{tabular}  \\ \hline
	\multirow{2}{*}{9}  & AUC   & \textbf{\begin{tabular}[c]{@{}c@{}}0.7154\\ (0.6918-0.7391)\end{tabular}}  & \begin{tabular}[c]{@{}c@{}}$0.7144^+$\\ (0.6886-0.7403)\end{tabular}  & \begin{tabular}[c]{@{}c@{}}$0.7069^+$\\ (0.6791-0.7347)\end{tabular} & \begin{tabular}[c]{@{}c@{}}$0.6926^*$\\ (0.6548-0.7305)\end{tabular} & \begin{tabular}[c]{@{}c@{}}$0.7130^{++}$\\ (0.6935-0.7325)\end{tabular}          & \begin{tabular}[c]{@{}c@{}}$0.6915^*$\\ (0.6487-0.7344)\end{tabular} & \begin{tabular}[c]{@{}c@{}}$0.6746^*$\\ (0.6463-0.7029)\end{tabular}  \\ \cline{2-9} 
	& KS    & \textbf{\begin{tabular}[c]{@{}c@{}}0.3048\\ (0.2835-0.3261)\end{tabular}}  & \begin{tabular}[c]{@{}c@{}}$0.3044^+$\\ (0.2812-0.3276)\end{tabular}  & \begin{tabular}[c]{@{}c@{}}$0.2893^+$\\ (0.2643-0.3143)\end{tabular} & \begin{tabular}[c]{@{}c@{}}$0.2719^*$\\ (0.2378-0.3059)\end{tabular} & \begin{tabular}[c]{@{}c@{}}$0.2988^{++}$\\ (0.2813-0.3164)\end{tabular}          & \begin{tabular}[c]{@{}c@{}}$0.2974^*$\\ (0.2588-0.3359)\end{tabular} & \begin{tabular}[c]{@{}c@{}}$0.2256^*$\\ (0.2001-0.2511)\end{tabular}  \\ \hline
	\multirow{2}{*}{11} & AUC   & \textbf{\begin{tabular}[c]{@{}c@{}}0.7161\\ (0.6927-0.7395)\end{tabular}}  & \begin{tabular}[c]{@{}c@{}}0.7146\\ (0.6891-0.7402)\end{tabular}  & \begin{tabular}[c]{@{}c@{}}$0.7079^+$\\ (0.6804-0.7354)\end{tabular} & \begin{tabular}[c]{@{}c@{}}$0.6934^*$\\ (0.6559-0.7309)\end{tabular} & \begin{tabular}[c]{@{}c@{}}0.7122\\ (0.6928-0.7315)\end{tabular}          & \begin{tabular}[c]{@{}c@{}}$0.6954^*$\\ (0.6530-0.7378)\end{tabular} & \begin{tabular}[c]{@{}c@{}}$0.6736^*$\\ (0.6455-0.7016)\end{tabular}  \\ \cline{2-9} 
	& KS    & \textbf{\begin{tabular}[c]{@{}c@{}}0.3052\\ (0.2846-0.3258)\end{tabular}}  & \begin{tabular}[c]{@{}c@{}}0.3031\\ (0.2807-0.3256)\end{tabular}  & \begin{tabular}[c]{@{}c@{}}$0.2913^+$\\ (0.2671-0.3155)\end{tabular} & \begin{tabular}[c]{@{}c@{}}$0.2771^*$\\ (0.2441-0.3101)\end{tabular} & \begin{tabular}[c]{@{}c@{}}0.3009\\ (0.2839-0.3179)\end{tabular}          & \begin{tabular}[c]{@{}c@{}}$0.2935^*$\\ (0.2562-0.3308)\end{tabular} & \begin{tabular}[c]{@{}c@{}}$0.2240^*$\\ (0.1993-0.2487)\end{tabular}  \\ \hline
	\multirow{2}{*}{13} & AUC   & \textbf{\begin{tabular}[c]{@{}c@{}}0.7139\\ (0.6920-0.7357)\end{tabular}}  & \begin{tabular}[c]{@{}c@{}}$0.7123^*$\\ (0.6885-0.7362)\end{tabular}  & \begin{tabular}[c]{@{}c@{}}$0.7063^*$\\ (0.6806-0.7320)\end{tabular} & \begin{tabular}[c]{@{}c@{}}$0.6907^*$\\ (0.6558-0.7256)\end{tabular} & \begin{tabular}[c]{@{}c@{}}$0.7096^*$\\ (0.6915-0.72766)\end{tabular}         & \begin{tabular}[c]{@{}c@{}}$0.6913^*$\\ (0.6517-0.7309)\end{tabular} & \begin{tabular}[c]{@{}c@{}}$0.6721^*$\\ (0.6460-0.6983)\end{tabular}  \\ \cline{2-9} 
	& KS    & \textbf{\begin{tabular}[c]{@{}c@{}}0.3009\\ (0.2821-0.3196)\end{tabular}}  & \begin{tabular}[c]{@{}c@{}}$0.2984^*$\\ (0.2779-0.3189)\end{tabular}  & \begin{tabular}[c]{@{}c@{}}$0.2888^*$\\ (0.2666-0.3109)\end{tabular} & \begin{tabular}[c]{@{}c@{}}$0.2664^*$\\ (0.2363-0.2964)\end{tabular} & \begin{tabular}[c]{@{}c@{}}$0.2980^*$\\ (0.2824-0.3135)\end{tabular}          & \begin{tabular}[c]{@{}c@{}}$0.2823^*$\\ (0.2482-0.3163)\end{tabular} & \begin{tabular}[c]{@{}c@{}}$0.2220^*$\\ (0.1995-0.2445)\end{tabular}  \\ \hline
	\multirow{2}{*}{15} & AUC   & \textbf{\begin{tabular}[c]{@{}c@{}}0.7081\\ (0.6889-0.7273)\end{tabular}}  & \begin{tabular}[c]{@{}c@{}}0.7048\\ (0.6839-0.7257)\end{tabular}  & \begin{tabular}[c]{@{}c@{}}$0.6988^+$\\ (0.6763-0.7213)\end{tabular} & \begin{tabular}[c]{@{}c@{}}$0.6859^*$\\ (0.6553-0.7166)\end{tabular} & \begin{tabular}[c]{@{}c@{}}$0.7018^+$\\ (0.6860-0.7176)\end{tabular}          & \begin{tabular}[c]{@{}c@{}}$0.6855^*$\\ (0.6508-0.7202)\end{tabular} & \begin{tabular}[c]{@{}c@{}}$0.6692^*$\\ (0.6462-0.6921)\end{tabular}  \\ \cline{2-9} 
	& KS    & \textbf{\begin{tabular}[c]{@{}c@{}}0.2937\\ (0.2775-0.3100)\end{tabular}}  & \begin{tabular}[c]{@{}c@{}}$0.2908^{++}$\\ (0.2730-0.3086)\end{tabular}  & \begin{tabular}[c]{@{}c@{}}$0.2798^+$\\ (0.2606-0.2989)\end{tabular} & \begin{tabular}[c]{@{}c@{}}$0.2583^*$\\ (0.2323-0.2844)\end{tabular} & \begin{tabular}[c]{@{}c@{}}$0.2871^+$\\ (0.2736-0.3005)\end{tabular}          & \begin{tabular}[c]{@{}c@{}}$0.2794^*$\\ (0.2499-0.3090)\end{tabular} & \begin{tabular}[c]{@{}c@{}}$0.2196^*$\\ (0.2001-0.2391)\end{tabular}  \\ \hline
	\multirow{2}{*}{17} & AUC   & \textbf{\begin{tabular}[c]{@{}c@{}}0.7029\\ (0.6867-0.7192)\end{tabular}}  & \begin{tabular}[c]{@{}c@{}}$0.7001^+$\\ (0.6823-0.7179)\end{tabular}  & \begin{tabular}[c]{@{}c@{}}$0.6948^*$\\ (0.6756-0.7139)\end{tabular} & \begin{tabular}[c]{@{}c@{}}$0.6817^*$\\ (0.6557-0.7078)\end{tabular} & \begin{tabular}[c]{@{}c@{}}$0.6972^*$\\ (0.6837-0.7106)\end{tabular}          & \begin{tabular}[c]{@{}c@{}}$0.6828^*$\\ (0.6533-0.7124)\end{tabular} & \begin{tabular}[c]{@{}c@{}}$0.6914^*$\\ (0.6719-0.7109)\end{tabular}  \\ \cline{2-9} 
	& KS    & \textbf{\begin{tabular}[c]{@{}c@{}}0.2857\\ (0.2721-0.2994)\end{tabular}}  & \begin{tabular}[c]{@{}c@{}}$0.2851^+$\\ (0.2702-0.3001)\end{tabular}  & \begin{tabular}[c]{@{}c@{}}$0.2758^*$\\ (0.2598-0.2919)\end{tabular} & \begin{tabular}[c]{@{}c@{}}$0.2552^*$\\ (0.2334-0.2771)\end{tabular} & \begin{tabular}[c]{@{}c@{}}$0.2835^*$\\ (0.2722-0.2948)\end{tabular}          & \begin{tabular}[c]{@{}c@{}}$0.2808^*$\\ (0.2561-0.3056)\end{tabular} & \begin{tabular}[c]{@{}c@{}}$0.2559^*$\\ (0.2395-0.2723)\end{tabular}  \\ \hline
	\multirow{2}{*}{19} & AUC   & \textbf{\begin{tabular}[c]{@{}c@{}}0.6966\\ (0.6836-0.7096)\end{tabular}}  & \begin{tabular}[c]{@{}c@{}}0.6930\\ (0.6788-0.7072)\end{tabular}  & \begin{tabular}[c]{@{}c@{}}$0.6885^+$\\ (0.6732-0.7038)\end{tabular} & \begin{tabular}[c]{@{}c@{}}$0.6756^*$\\ (0.6548-0.6964)\end{tabular} & \begin{tabular}[c]{@{}c@{}}$0.6898^*$\\ (0.67901-0.7006)\end{tabular}         & \begin{tabular}[c]{@{}c@{}}$0.6772^*$\\ (0.6536-0.7007)\end{tabular} & \begin{tabular}[c]{@{}c@{}}$0.6823^+$\\ (0.6667-0.6979)\end{tabular}  \\ \cline{2-9} 
	& KS    & \textbf{\begin{tabular}[c]{@{}c@{}}0.2771\\ (0.2666-0.2876)\end{tabular}}  & \begin{tabular}[c]{@{}c@{}}0.2752\\ (0.2637-0.2867)\end{tabular}  & \begin{tabular}[c]{@{}c@{}}$0.2681^+$\\ (0.2557-0.2804)\end{tabular} & \begin{tabular}[c]{@{}c@{}}$0.2467^*$\\ (0.2299-0.2636)\end{tabular} & \begin{tabular}[c]{@{}c@{}}$0.2753^*$\\ (0.2666-0.2840)\end{tabular}          & \begin{tabular}[c]{@{}c@{}}$0.2656^*$\\ (0.2465-0.2847)\end{tabular} & \begin{tabular}[c]{@{}c@{}}$0.2471^+$\\ (0.2345-0.2597)\end{tabular} \\ \hline
	\multirow{2}{*}{21} & AUC   & \textbf{\begin{tabular}[c]{@{}c@{}}0.6911\\ (0.6814-0.7008)\end{tabular}}  & \begin{tabular}[c]{@{}c@{}}$0.6868^+$\\ (0.6763-0.6974)\end{tabular}  & \begin{tabular}[c]{@{}c@{}}$0.6835^+$\\ (0.6721-0.6949)\end{tabular} & \begin{tabular}[c]{@{}c@{}}$0.6715^*$\\ (0.6560-0.6870)\end{tabular} & \begin{tabular}[c]{@{}c@{}}0.6839\\ (0.6759-0.6919)\end{tabular}          & \begin{tabular}[c]{@{}c@{}}$0.6747^*$\\ (0.6572-0.6923)\end{tabular} & \begin{tabular}[c]{@{}c@{}}$0.6792^*$\\ (0.6676-0.6908)\end{tabular}  \\ \cline{2-9} 
	& KS    & \textbf{\begin{tabular}[c]{@{}c@{}}0.2716\\ (0.2639-0.2792)\end{tabular}}  & \begin{tabular}[c]{@{}c@{}}$0.2691^+$\\ (0.2607-0.2774)\end{tabular}  & \begin{tabular}[c]{@{}c@{}}$0.2635^+$\\ (0.2545-0.2725)\end{tabular} & \begin{tabular}[c]{@{}c@{}}$0.2413^*$\\ (0.2290-0.2535)\end{tabular} & \begin{tabular}[c]{@{}c@{}}0.2687\\ (0.2624-0.2750)\end{tabular}          & \begin{tabular}[c]{@{}c@{}}$0.2622^*$\\ (0.2483-0.2760)\end{tabular} & \begin{tabular}[c]{@{}c@{}}$0.2439^*$\\ (0.2347-0.2530)\end{tabular}  \\ \hline
	\multirow{2}{*}{23} & AUC   & \textbf{\begin{tabular}[c]{@{}c@{}}0.6875\\ (0.6806-0.6944)\end{tabular}}  & \begin{tabular}[c]{@{}c@{}}$0.6829^{++}$\\ (0.6753-0.6904)\end{tabular}  & \begin{tabular}[c]{@{}c@{}}$0.6797^+$\\ (0.6716-0.6878)\end{tabular} & \begin{tabular}[c]{@{}c@{}}$0.6671^*$\\ (0.6561-0.6781)\end{tabular} & \begin{tabular}[c]{@{}c@{}}$0.6796^{++}$\\ (0.6739-0.6853)\end{tabular}          & \begin{tabular}[c]{@{}c@{}}$0.6744^*$\\ (0.6619-0.6869)\end{tabular} & \begin{tabular}[c]{@{}c@{}}$0.6576^*$\\ (0.6493-0.6658)\end{tabular}  \\ \cline{2-9} 
	& KS    & \textbf{\begin{tabular}[c]{@{}c@{}}0.2668\\ (0.26159-0.2721)\end{tabular}} & \begin{tabular}[c]{@{}c@{}}$0.2662^{++}$\\ (0.2605-0.2719)\end{tabular}  & \begin{tabular}[c]{@{}c@{}}$0.2599^+$\\ (0.2538-0.2661)\end{tabular} & \begin{tabular}[c]{@{}c@{}}$0.2364^*$\\ (0.2281-0.2448)\end{tabular} & \begin{tabular}[c]{@{}c@{}}$0.2640^{++}$\\ (0.2597-0.2683)\end{tabular}          & \begin{tabular}[c]{@{}c@{}}$0.2660^*$\\ (0.2565-0.2755)\end{tabular} & \begin{tabular}[c]{@{}c@{}}$0.2173^*$\\ (0.2110-0.2235)\end{tabular}  \\ \hline
\end{tabular}}
\begin{tablenotes}
	\footnotesize
	\item[$*$] indicates p-value $ < 0.001$
	\item[$+$] indicates p-value $ < 0.01$
	\item[$++$] indicates p-value $ < 0.05$
\end{tablenotes}
\end{threeparttable}
\caption{AUC and KS indices of all candidate models (in terms of mean values and 95\% confidence intervals) for every 2 repayment months. The highest score of each row is highlighted in boldface type.}
\label{table:auc_ks}
\end{table}
}

Table \ref{table:auc_ks} presents the results of AUC and KS index for every 2 repayment months by all candidate models except GBMCI. For each model, the first observation is that except for a few cases, the AUC (respectively, KS index) decreases, and hence, the discriminating power deceases, as time (month) increases. It can be explained by the fact that the prediction of future events is more difficult than that of recent events. The second observation from Table \ref{table:auc_ks} is that except for the 3rd month, GBST achieves the highest score (in terms of mean value) among all models for all months in terms of both performance metrics, and the difference between GBST and the other models are statistically significant in terms of p-value ($<0.05$) with a few exceptions. Therefore, we conclude that GBST performs significantly better than all other models over all months (with one exception) in the lending club loan dataset.

\begin{figure}[!ht]
  \centering
    \includegraphics[width=0.8\textwidth]{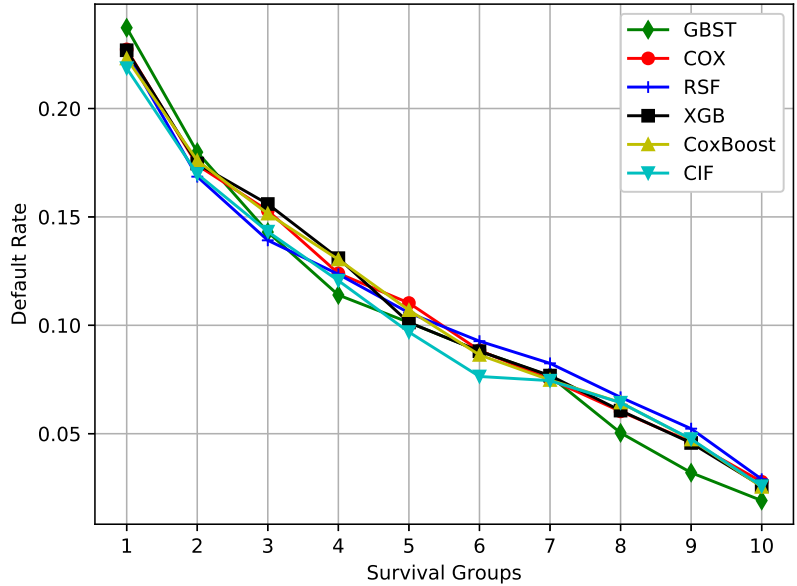}
   \caption{Default rate curves of different survival groups in the test dataset (May 2013 - Dec.\ 2013, see Table \ref{table:lcldata}) in 24th repayment month. For each model, loans in the test dataset are sorted in ascending order by the predicted survival probabilities and are then divided into 10 groups with equal size. The larger the group index is, the higher the predicted survival probabilities of the group are.} 
   \label{fig:default3lc}
\end{figure}

Finally, in order to understand the performance variation in the models of interest, we conduct an analysis similar to the one presented in Section \ref{sec:performance} by comparing the predicted survival probabilities with the true default rate. For each model, the loans are sorted in ascending order by the predicted survival probability of the 24th repayment month and are accordingly divided into 10 survival groups with equal sample size. It is noteworthy that since for each loan the survival probabilities predicted by different models may differ from each other, a loan can be categorized into different survival groups under different models. For each survival group predicted by each model, we calculate the default rate of the 24th month within the group. The results are plotted in Figure \ref{fig:default3lc}. We find that first, all models shows to some extent a discriminating power, since the lower survival group index is (i.e., the lower the average predicted probability within the group is), the higher the default rate is; and second, GBST shows a better discriminating power than the other models, as for the bottom 4 survival groups (1 -- 4) predicted by GBST have higher default rates than those groups by other models, while the top 4 survival groups (7 -- 10) predicted by GBST have lower default rates than those groups by other models.


To conclude this subsection, the proposed GBST model outperforms the other candidate models (COX, RSF, XGB, CoxBoost, GBMCI, CIF and DeepHit) on the lending club loan dataset, measured by three typical performance metrics (C-index, AUC, KS index), as well as by the group analysis conducted above.

\subsection{360 Finance Dataset}


\subsubsection{Dataset}

We use the data provided by 360 Finance Inc., who provides credit products mainly in form of installment loans. The loans are expected to be repaid monthly over a period of 3, 6 or 12 months. In order to make a full observation of repayment behavior, we consider only the accounts who borrowed money in January 2018 and March 2018. The loans issued in February 2018 are excluded to avoid the festival effect for the Chinese New Year Festival was in that month. All the loans were scheduled to be repaid in 12-month installments. 
Among them, 200,000 loans borrowed in January are randomly selected as the training set, and 120,000 in March are sampled as the test set (see Table \ref{table:data}). In total, 320,000 samples are drawn.

\begin{figure}[!ht]
  \centering
    \includegraphics[width=0.8\textwidth]{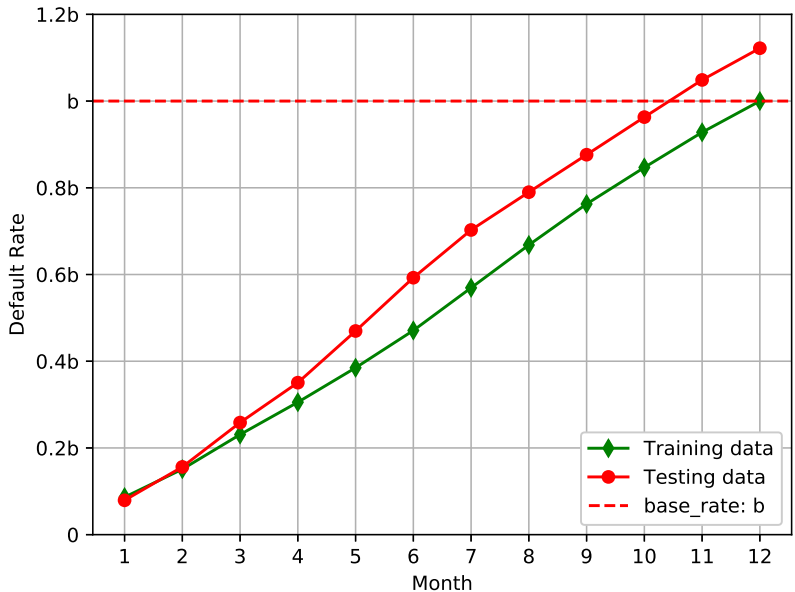}
   \caption{(Cumulative) default rates (see Eq.\ \eqref{eq:dr}) for both training and test data, compared to the baseline default rate $b$, which is the default rate of the sample accounts from the training dataset cummulated to 12th month. For instance, $0.2b$ means that the default rate is 20\% of the default rate $b$.}
   \label{fig:default}
\end{figure}

A loan is defined as a \emph{default} loan in this dataset if on any scheduled repayment due date the borrower is once overdue for at least 10 days, no matter whether he pays off the installment later or not. Hence, the discretized observation time (see Section \ref{subsection:sa}) applied in survival analysis is the monthly repayment due date with extra 10 days. Therefore, the \emph{default rate} in certain repayment month is the number of accounts who is overdue for at least 10 days in this month, as depicted in Figure \ref{fig:default}. In compliance with the data protection policy of 360 Finance Inc., the exact values of default rates are not plotted. Instead, we plot the values of default rates compared to a baseline, which is the default rate of the sample accounts from the training dataset cumulated to 12th month. It should also be noted that our definition of default is stricter than the conventional definition and therefore, the default rate in our definition is higher as well. We also see in Figure \ref{fig:default} that the default rate of the test data in March 2018 grows faster than that of the training data in January 2018, though two curves do not diverge significantly.


\begin{table}[!ht]
\centering
\begin{tabular}{c|c|c}
\hline
{\bf dataset}& \bf transaction date& \bf sample size\\
\hline
training set& January 2018 & 200,000\\
test set& March 2018 & 120,000\\
\hline
\end{tabular}
\caption{Summary of datasets.}
\label{table:data}
\end{table}

\subsubsection{Pre-processing of data and feature selection}
\label{sec:preprocessing2}

For each loan, over 400 raw variables are collected at application, including
\begin{itemize}
 \item personal information provided by the borrower, e.g., sex, age, education level;
 \item credit report retrieved from People's Bank of China (PBC), e.g., income score, credit score;
 \item device information collected by the 360 Finance App which is authorized by the borrower, e.g., device brand, memory size of the mobile phone;
 \item credit report from third-party credit rate agencies, e.g., number of loans applied in other platforms, travel intensity, etc.
\end{itemize}

We conduct a pre-processing procedure similar to the one described in Section \ref{sec:preprocessing}. The variables with a missing rate higher than 80\% are excluded and the categorical variables are transformed into numerical values with one-hot encoding. We fill up the missing values in each attribute with a default value and add a missing-value indicator. After pre-processing, ca.\ 1000 features are generated. 

Although GBST and XGB can deal with high dimensional inputs as in this case more than 1000, all of the other models to be compared with can only work with inputs of a rather smaller dimensionality, which is usually less than 100. Hence, for the sake of fair comparison, we apply XGB to do a feature selection before training by survival models. More specifically, we train an XGB model with the training dataset to solve a classification problem, where the samples are labeled as ``bad'' if the account defaults within the observation period of 12 months, and are labeled as ``good'' otherwise. The hyperparameters of XGB model, such as max tree depth, are chosen based on a 5-fold cross validation. We finally select the 50 features with the highest importance (for details see Section 10.13, \citealp{friedman2001elements}). Table \ref{table:features} shows some of the most important features and their data sources.


\begin{table}[!ht]
 \centering
 \begin{tabular}{|c|c|}
  \hline
  source & feature \\
  \hline

  \multirow{3}{*}{PBC report} & income score\\
  \cline{2-2}
  & credit score \\
  \cline{2-2}
  & overdue information\ of credit cards\\
  \hline

  \multirow{3}{*}{personal information} & age\\
  \cline{2-2}
  & sex \\
  \cline{2-2}
  & education level\\
  \hline
  \multirow{2}{*}{third-party rate agency} & no.\ of loans in other lending platforms\\
  \cline{2-2}
  & travel intensity \\
  \hline
  \multirow{2}{*}{other information} & whether possessing a car\\
  \cline{2-2}
  & application channel\\
  \hline
 \end{tabular}
 \caption{List of important features, which are selected according to the feature importance generated by an XGB model trained with the training dataset.}
\label{table:features}
\end{table}

\subsubsection{Performance and comparison}
\label{sec:perf}
In this subsection we follow the same steps of experiments done for the lending club loan dataset described in Subsection \ref{sec:lcld}. 

With 50 features selected XGB as inputs, we train GBST model on the training set of Jan.\ 2018 and apply the obtained model to the test set of Mar.\ 2018. For each repayment month, the loans are sorted in ascending order according to the probabilities predicted by GBST and are then divided into 10 groups with equal sample size. Therefore, the 1st group contains the loans with the lowest predicted survival probability, i.e., the highest predicted default rate, and vice versa. Figure \ref{fig:default2} exhibits the plots of the true default rate of each group across the whole period of loan. The primary findings are as follows:

\begin{figure}[!ht]
  \centering
    \includegraphics[width=0.8\textwidth]{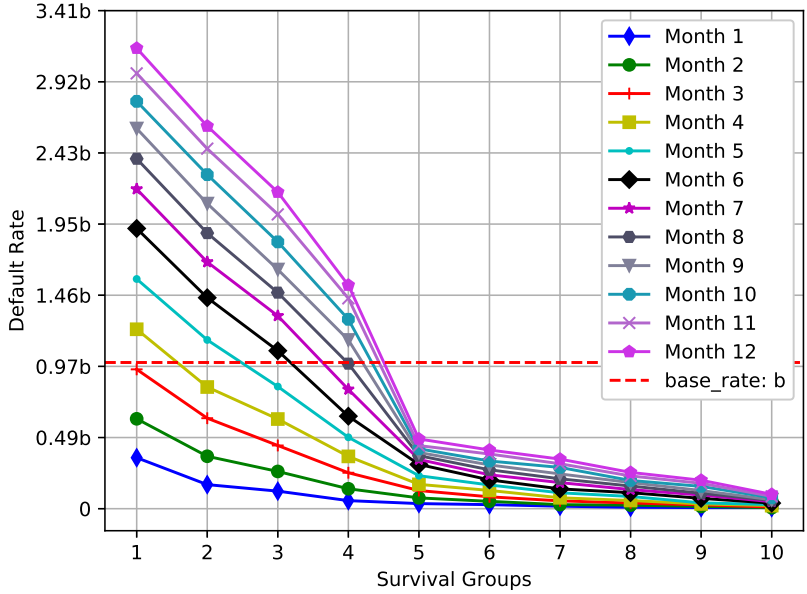}
   \caption{Default rate curves of different survival groups in the test dataset over repayment months. Accounts in the test dataset (March 2018) are sorted in ascending order by the survival probability predicted by GBST.}
   \label{fig:default2}
\end{figure}

\begin{itemize}
 \item The predicted survival probability is negatively related to the true default rate in all the repayment months, i.e., groups with higher (lower) predicted survival probability generally have lower (higher) real default rates.
\item The slope of the real default rate line becomes much steeper in the interval between the 4th group and the 5th group, which implies that the survival probability in this interval has stronger discriminating power of default.
\end{itemize}

We then compare the performance with other models. We conduct a 10 leave-one-out cross-validation (LOOCV) experiments with 10 almost equal-size datasets (see Table \ref{table:data2}). Similar to the training procedure described in Subsection \ref{sec:lcld}, in each experiment, the hyper\-parameters of each model are optimized by a grid search with 5-fold cross validation on the training set. 


\begin{table}[!ht]
	\centering
	\begin{tabular}{c|c|c}
		\hline
		{\bf index}& \bf first transaction date& \bf sample size\\
		\hline
		1 & January 1, 2018 to January 4, 2018 & 34,904\\
		2 & January 5, 2018 to January 8, 2018 & 30,355\\
		3 & January 9, 2018 to January 13, 2018 & 34,868\\
		4 & January 14, 2018 to January 17, 2018 & 27,143\\
		5 & January 18, 2018 to January 23, 2018 & 35,978\\
		6 & January 24, 2018 to January 28, 2018 & 27,309\\
		7 & January 29, 2018 to March 3, 2018 & 35,273\\
		8 & March 4, 2018 to March 8, 2018& 34,163\\
		9 & March 9, 2018 to March 13, 2018 & 30,273\\
		10 & March 14, 2018 to March 20, 2018 & 29,734\\
		\hline
	\end{tabular}
	\caption{10 datasets of 360 Finance data by date of first transaction (the date when loans are borrowed).}
	\label{table:data2}
\end{table}

We calculate the C-index to measure the performance of all candidate models in the 360 Finance dataset and the results are presented in Table \ref{table:cindex_n}. It shows that GBST achieves the highest score (0.7997) among all models and the difference between its score and other scores is significantly large in terms of p-value ($< 0.05$). Hence, we can conclude that GBST outperforms all the other models significantly and it has a strong discriminating power for risk assessment.

Furthermore, comparing the results shown in Table \ref{table:cindex_n} with those in Table \ref{table:cindex_n2}, we find that the scores of all models on this dataset are much higher than the scores on the lending club loan dataset (cf.\ Table \ref{table:cindex_n2}). It suggests that the learning task on the 360 Finance data is easier than the task on the lending club loan data, probably because the 360 Finance dataset contains not only more variables than the lending club loan dataset (400 vs.\ 50), but the variables in the 360 Finance dataset are also more relevant to the risk assessment than the variables in the latter dataset. 


 
\setlength{\tabcolsep}{2mm}{
	\begin{table}[]
		\centering
		\begin{threeparttable}
			\begin{tabular}{|c|c|c|c|c|}
				\hline
				Algorithm & GBST                                                                           & COX                                                                   & RSF                                                                   & XGB                                                                   \\ \hline
				C-index   & \textbf{\begin{tabular}[c]{@{}c@{}}0.7997\\      (0.7826-0.8167)\end{tabular}} & \begin{tabular}[c]{@{}c@{}}$0.7797^+$\\      (0.7652-0.7941)\end{tabular} & \begin{tabular}[c]{@{}c@{}}$0.7765^*$\\      (0.7576-0.7955)\end{tabular} & \begin{tabular}[c]{@{}c@{}}$0.7742^*$\\      (0.7537-0.7947)\end{tabular} \\ \hline
				Algorithm & CoxBoost                                                                       & GBMCI                                                                 & CIF                                                                   & DeepHit                                                               \\ \hline
				C-index   & \begin{tabular}[c]{@{}c@{}}$0.7691^{+}$\\      (0.7529-0.7853)\end{tabular}          & \begin{tabular}[c]{@{}c@{}}$0.6905^*$\\      (0.6767-0.7043)\end{tabular} & \begin{tabular}[c]{@{}c@{}}$0.7395^*$\\      (0.7189-0.7601)\end{tabular} & \begin{tabular}[c]{@{}c@{}}$0.7808^{++}$\\      (0.7659-0.7957)\end{tabular} \\ \hline
			\end{tabular}
			\begin{tablenotes}
				\footnotesize
				\item[$*$] indicates p-value $ < 0.001$
				\item[$+$] indicates p-value $ < 0.01$
				\item[$++$] indicates p-value $ < 0.05$
			\end{tablenotes}
		\end{threeparttable}
		\caption{C-indices of 10 LOOCV experiments (in terms of mean values and 95\% confidence intervals).}
		\label{table:cindex_n}
	\end{table}
}



We also calculate AUC and KS indices for each candidate model and each repayment month. The results are summarized in Table \ref{tab:auc_ks_2}. It shows that both AUC and KS indices of GBST are higher than the corresponding scores of the other models and the differences are statistically significant (p-value $< 0.05$) with very few exceptions. Hence, GBST outperforms the other models significantly. Furthermore, both scores of GBST decays much more slowly with time than the scores of the other models, indicating that the GBST model tends to be more time-consistent on this dataset.

\setlength{\tabcolsep}{7mm}{
\begin{table}[]
\centering
\begin{threeparttable}
\resizebox{\textwidth}{!}{%
\begin{tabular}{|c|c|c|c|c|c|c|c|c|}
	\hline
	Month               & Index & GBST                                                                           & XGB                                                                    & COX                                                                   & RSF                                                                    & CoxBoost                                                              & CIF                                                                   & DeepHit                                                               \\ \hline
	\multirow{2}{*}{1}  & AUC   & \textbf{\begin{tabular}[c]{@{}c@{}}0.8251\\      (0.8046-0.8456)\end{tabular}} & \begin{tabular}[c]{@{}c@{}}$0.8068^+$\\      (0.78945-0.8241)\end{tabular} & \begin{tabular}[c]{@{}c@{}}$0.8124^*$\\      (0.7897-0.8352)\end{tabular} & \begin{tabular}[c]{@{}c@{}}$0.8180^+$\\      (0.7934-0.8426)\end{tabular}  & \begin{tabular}[c]{@{}c@{}}$0.7976^*$\\      (0.7781-0.8170)\end{tabular} & \begin{tabular}[c]{@{}c@{}}$0.6959^*$\\      (0.6711-0.7206)\end{tabular} & \begin{tabular}[c]{@{}c@{}}$0.7925^*$\\      (0.7746-0.8104)\end{tabular} \\ \cline{2-9} 
	& KS    & \textbf{\begin{tabular}[c]{@{}c@{}}0.5164\\      (0.4959-0.5369)\end{tabular}} & \begin{tabular}[c]{@{}c@{}}$0.4687^+$\\      (0.4513-0.4860)\end{tabular}  & \begin{tabular}[c]{@{}c@{}}$0.4766^*$\\      (0.4539-0.4994)\end{tabular} & \begin{tabular}[c]{@{}c@{}}$0.4999^+$\\      (0.4753-0.5245)\end{tabular}  & \begin{tabular}[c]{@{}c@{}}$0.4561^*$\\      (0.4366-0.4755)\end{tabular} & \begin{tabular}[c]{@{}c@{}}$0.3377^*$\\      (0.3129-0.3624)\end{tabular} & \begin{tabular}[c]{@{}c@{}}$0.4608^*$\\      (0.4429-0.4786)\end{tabular} \\ \hline
	\multirow{2}{*}{2}  & AUC   & \textbf{\begin{tabular}[c]{@{}c@{}}0.8225\\      (0.7984-0.8467)\end{tabular}} & \begin{tabular}[c]{@{}c@{}}$0.8060^*$\\      (0.7855-0.8265)\end{tabular}  & \begin{tabular}[c]{@{}c@{}}$0.8149^+$\\      (0.7881-0.8417)\end{tabular} & \begin{tabular}[c]{@{}c@{}}$0.8149^*$\\      (0.7859-0.8439)\end{tabular}  & \begin{tabular}[c]{@{}c@{}}$0.7952^{++}$\\      (0.7723-0.8181)\end{tabular} & \begin{tabular}[c]{@{}c@{}}$0.7609^*$\\      (0.7317-0.7900)\end{tabular} & \begin{tabular}[c]{@{}c@{}}$0.7906^*$\\      (0.7695-0.8117)\end{tabular} \\ \cline{2-9} 
	& KS    & \textbf{\begin{tabular}[c]{@{}c@{}}0.5135\\      (0.4899-0.5372)\end{tabular}} & \begin{tabular}[c]{@{}c@{}}$0.4616^*$\\      (0.4416-0.4817)\end{tabular}  & \begin{tabular}[c]{@{}c@{}}$0.4775^+$\\      (0.4512-0.5038)\end{tabular} & \begin{tabular}[c]{@{}c@{}}$0.4944^*$\\      (0.4660-0.5229)\end{tabular}  & \begin{tabular}[c]{@{}c@{}}$0.4480^{++}$\\      (0.4255-0.4705)\end{tabular} & \begin{tabular}[c]{@{}c@{}}$0.4043^*$\\      (0.3757-0.4329)\end{tabular} & \begin{tabular}[c]{@{}c@{}}$0.4542^+$\\      (0.4335-0.4749)\end{tabular} \\ \hline
	\multirow{2}{*}{3}  & AUC   & \textbf{\begin{tabular}[c]{@{}c@{}}0.8224\\      (0.7950-0.8498)\end{tabular}} & \begin{tabular}[c]{@{}c@{}}$0.8048^+$\\      (0.7815-0.8280)\end{tabular}  & \begin{tabular}[c]{@{}c@{}}$0.8120^*$\\      (0.7815-0.8425)\end{tabular} & \begin{tabular}[c]{@{}c@{}}$0.8158^{++}$\\      (0.7828-0.8488)\end{tabular}  & \begin{tabular}[c]{@{}c@{}}$0.7975^*$\\      (0.7714-0.8236)\end{tabular} & \begin{tabular}[c]{@{}c@{}}$0.7650^*$\\      (0.7319-0.7982)\end{tabular} & \begin{tabular}[c]{@{}c@{}}$0.7923^*$\\      (0.7683-0.8163)\end{tabular} \\ \cline{2-9} 
	& KS    & \textbf{\begin{tabular}[c]{@{}c@{}}0.5145\\      (0.4887-0.5403)\end{tabular}} & \begin{tabular}[c]{@{}c@{}}$0.4556^+$\\      (0.4338-0.4775)\end{tabular}  & \begin{tabular}[c]{@{}c@{}}$0.4707^*$\\      (0.4420-0.4993)\end{tabular} & \begin{tabular}[c]{@{}c@{}}$0.4943^{++}$\\      (0.4632-0.52538)\end{tabular} & \begin{tabular}[c]{@{}c@{}}$0.4495^*$\\      (0.4250-0.4740)\end{tabular} & \begin{tabular}[c]{@{}c@{}}$0.4060^*$\\      (0.3749-0.4372)\end{tabular} & \begin{tabular}[c]{@{}c@{}}$0.4551^*$\\      (0.4325-0.4776)\end{tabular} \\ \hline
	\multirow{2}{*}{4}  & AUC   & \textbf{\begin{tabular}[c]{@{}c@{}}0.8199\\      (0.7902-0.8496)\end{tabular}} & \begin{tabular}[c]{@{}c@{}}$0.8051^*$\\      (0.7800-0.8303)\end{tabular}  & \begin{tabular}[c]{@{}c@{}}0.8105\\      (0.7775-0.8434)\end{tabular} & \begin{tabular}[c]{@{}c@{}}0.8133\\      (0.7776-0.8489)\end{tabular}  & \begin{tabular}[c]{@{}c@{}}$0.7973^+$\\      (0.7692-0.8255)\end{tabular} & \begin{tabular}[c]{@{}c@{}}$0.7488^*$\\      (0.7130-0.7847)\end{tabular} & \begin{tabular}[c]{@{}c@{}}$0.7921^*$\\      (0.7662-0.8180)\end{tabular} \\ \cline{2-9} 
	& KS    & \textbf{\begin{tabular}[c]{@{}c@{}}0.5098\\      (0.4825-0.5371)\end{tabular}} & \begin{tabular}[c]{@{}c@{}}$0.4571^*$\\      (0.4340-0.4802)\end{tabular}  & \begin{tabular}[c]{@{}c@{}}0.4653\\      (0.4350-0.4957)\end{tabular} & \begin{tabular}[c]{@{}c@{}}0.4954\\      (0.4626-0.5283)\end{tabular}  & \begin{tabular}[c]{@{}c@{}}$0.4467^+$\\      (0.4207-0.4726)\end{tabular} & \begin{tabular}[c]{@{}c@{}}$0.379^*6$\\      (0.3467-0.4126)\end{tabular} & \begin{tabular}[c]{@{}c@{}}$0.4476^*$\\      (0.4238-0.4715)\end{tabular} \\ \hline
	\multirow{2}{*}{5}  & AUC   & \textbf{\begin{tabular}[c]{@{}c@{}}0.8210\\      (0.7904-0.8515)\end{tabular}} & \begin{tabular}[c]{@{}c@{}}$0.8041^{++}$\\      (0.7782-0.8300)\end{tabular}  & \begin{tabular}[c]{@{}c@{}}$0.8088^+$\\      (0.7749-0.8428)\end{tabular} & \begin{tabular}[c]{@{}c@{}}$0.8051^+$\\      (0.7684-0.8419)\end{tabular}  & \begin{tabular}[c]{@{}c@{}}$0.7967^*$\\      (0.7677-0.8258)\end{tabular} & \begin{tabular}[c]{@{}c@{}}$0.7420^*$\\      (0.7050-0.7789)\end{tabular} & \begin{tabular}[c]{@{}c@{}}$0.7903^*$\\      (0.7636-0.8170)\end{tabular} \\ \cline{2-9} 
	& KS    & \textbf{\begin{tabular}[c]{@{}c@{}}0.5107\\      (0.4832-0.5383)\end{tabular}} & \begin{tabular}[c]{@{}c@{}}$0.4557^{++}$\\      (0.4324-0.4790)\end{tabular}  & \begin{tabular}[c]{@{}c@{}}$0.4591^+$\\      (0.4285-0.4896)\end{tabular} & \begin{tabular}[c]{@{}c@{}}$0.4865^+$\\      (0.4535-0.5196)\end{tabular}  & \begin{tabular}[c]{@{}c@{}}$0.4426^*$\\      (0.4165-0.4688)\end{tabular} & \begin{tabular}[c]{@{}c@{}}$0.3703^*$\\      (0.3370-0.4035)\end{tabular} & \begin{tabular}[c]{@{}c@{}}$0.4403^*$\\      (0.4162-0.4643)\end{tabular} \\ \hline
	\multirow{2}{*}{6}  & AUC   & \textbf{\begin{tabular}[c]{@{}c@{}}0.8186\\      (0.7883-0.8488)\end{tabular}} & \begin{tabular}[c]{@{}c@{}}$0.7999^*$\\      (0.7742-0.8255)\end{tabular}  & \begin{tabular}[c]{@{}c@{}}0.8040\\      (0.7704-0.8377)\end{tabular} & \begin{tabular}[c]{@{}c@{}}$0.8021^+$\\      (0.7657-0.8385)\end{tabular}  & \begin{tabular}[c]{@{}c@{}}$0.7970^*$\\      (0.7683-0.8258)\end{tabular} & \begin{tabular}[c]{@{}c@{}}$0.7538^*$\\      (0.7172-0.7904)\end{tabular} & \begin{tabular}[c]{@{}c@{}}$0.7882^*$\\      (0.7618-0.8147)\end{tabular} \\ \cline{2-9} 
	& KS    & \textbf{\begin{tabular}[c]{@{}c@{}}0.5113\\      (0.4847-0.5379)\end{tabular}} & \begin{tabular}[c]{@{}c@{}}$0.4508^*$\\      (0.4282-0.4734)\end{tabular}  & \begin{tabular}[c]{@{}c@{}}0.4525\\      (0.4229-0.4821)\end{tabular} & \begin{tabular}[c]{@{}c@{}}$0.4862^+$\\      (0.4542-0.5182)\end{tabular}  & \begin{tabular}[c]{@{}c@{}}$0.4421^*$\\      (0.4168-0.4674)\end{tabular} & \begin{tabular}[c]{@{}c@{}}$0.3851^*$\\      (0.3530-0.4173)\end{tabular} & \begin{tabular}[c]{@{}c@{}}$0.4366^*$\\      (0.4133-0.4599)\end{tabular} \\ \hline
	\multirow{2}{*}{7}  & AUC   & \textbf{\begin{tabular}[c]{@{}c@{}}0.8156\\      (0.7874-0.8438)\end{tabular}} & \begin{tabular}[c]{@{}c@{}}$0.7944^*$\\      (0.7705-0.8184)\end{tabular}  & \begin{tabular}[c]{@{}c@{}}$0.7980^*$\\      (0.7667-0.8294)\end{tabular} & \begin{tabular}[c]{@{}c@{}}$0.7941^*$\\      (0.7602-0.8281)\end{tabular}  & \begin{tabular}[c]{@{}c@{}}$0.7921^*$\\      (0.7653-0.8189)\end{tabular} & \begin{tabular}[c]{@{}c@{}}$0.7660^*$\\      (0.7319-0.8001)\end{tabular} & \begin{tabular}[c]{@{}c@{}}$0.7883^*$\\      (0.7637-0.8130)\end{tabular} \\ \cline{2-9} 
	& KS    & \textbf{\begin{tabular}[c]{@{}c@{}}0.5116\\      (0.4873-0.5359)\end{tabular}} & \begin{tabular}[c]{@{}c@{}}$0.4411^*$\\      (0.4205-0.4616)\end{tabular}  & \begin{tabular}[c]{@{}c@{}}$0.4418^*$\\      (0.4148-0.4688)\end{tabular} & \begin{tabular}[c]{@{}c@{}}$0.4708^*$\\      (0.4416-0.5000)\end{tabular}  & \begin{tabular}[c]{@{}c@{}}$0.4317^*$\\      (0.4086-0.4548)\end{tabular} & \begin{tabular}[c]{@{}c@{}}$0.4111^*$\\      (0.3818-0.4405)\end{tabular} & \begin{tabular}[c]{@{}c@{}}$0.4363^*$\\      (0.4151-0.4576)\end{tabular} \\ \hline
	\multirow{2}{*}{8}  & AUC   & \textbf{\begin{tabular}[c]{@{}c@{}}0.8120\\      (0.7872-0.8368)\end{tabular}} & \begin{tabular}[c]{@{}c@{}}0.7900\\      (0.7691-0.8110)\end{tabular}  & \begin{tabular}[c]{@{}c@{}}$0.7932^*$\\      (0.7657-0.8207)\end{tabular} & \begin{tabular}[c]{@{}c@{}}$0.7904^*$\\      (0.7606-0.8202)\end{tabular}  & \begin{tabular}[c]{@{}c@{}}$0.7851^*$\\      (0.7616-0.8086)\end{tabular} & \begin{tabular}[c]{@{}c@{}}$0.7665^*$\\      (0.7366-0.7964)\end{tabular} & \begin{tabular}[c]{@{}c@{}}$0.7840^*$\\      (0.7624-0.8057)\end{tabular} \\ \cline{2-9} 
	& KS    & \textbf{\begin{tabular}[c]{@{}c@{}}0.5103\\      (0.4892-0.5313)\end{tabular}} & \begin{tabular}[c]{@{}c@{}}0.4349\\      (0.4171-0.4528)\end{tabular}  & \begin{tabular}[c]{@{}c@{}}$0.4350^*$\\      (0.4116-0.4584)\end{tabular} & \begin{tabular}[c]{@{}c@{}}$0.4686^*$\\      (0.4433-0.4939)\end{tabular}  & \begin{tabular}[c]{@{}c@{}}$0.4224^*$\\      (0.4024-0.4424)\end{tabular} & \begin{tabular}[c]{@{}c@{}}$0.4067^*$\\      (0.3813-0.4322)\end{tabular} & \begin{tabular}[c]{@{}c@{}}$0.4298^*$\\      (0.4114-0.4482)\end{tabular} \\ \hline
	\multirow{2}{*}{9}  & AUC   & \textbf{\begin{tabular}[c]{@{}c@{}}0.8095\\      (0.7884-0.8305)\end{tabular}} & \begin{tabular}[c]{@{}c@{}}$0.7863^{++}$\\      (0.7685-0.8042)\end{tabular}  & \begin{tabular}[c]{@{}c@{}}$0.7878^+$\\      (0.7644-0.8112)\end{tabular} & \begin{tabular}[c]{@{}c@{}}$0.7871^*$\\      (0.7618-0.8124)\end{tabular}  & \begin{tabular}[c]{@{}c@{}}$0.7806^*$\\      (0.7606-0.8006)\end{tabular} & \begin{tabular}[c]{@{}c@{}}$0.7639^*$\\      (0.7385-0.7894)\end{tabular} & \begin{tabular}[c]{@{}c@{}}$0.7804^+$\\      (0.7620-0.7988)\end{tabular} \\ \cline{2-9} 
	& KS    & \textbf{\begin{tabular}[c]{@{}c@{}}0.5093\\      (0.4917-0.5270)\end{tabular}} & \begin{tabular}[c]{@{}c@{}}$0.4301^{++}$\\      (0.4151-0.4451)\end{tabular}  & \begin{tabular}[c]{@{}c@{}}$0.4285^+$\\      (0.4089-0.4482)\end{tabular} & \begin{tabular}[c]{@{}c@{}}$0.4683^*$\\      (0.4471-0.4896)\end{tabular}  & \begin{tabular}[c]{@{}c@{}}$0.4177^*$\\      (0.4009-0.4345)\end{tabular} & \begin{tabular}[c]{@{}c@{}}$0.4078^+$\\      (0.3865-0.4292)\end{tabular} & \begin{tabular}[c]{@{}c@{}}$0.4274^*$\\      (0.4119-0.4428)\end{tabular} \\ \hline
	\multirow{2}{*}{10} & AUC   & \textbf{\begin{tabular}[c]{@{}c@{}}0.8061\\      (0.7893-0.8229)\end{tabular}} & \begin{tabular}[c]{@{}c@{}}$0.7808^+$\\      (0.7666-0.7950)\end{tabular}  & \begin{tabular}[c]{@{}c@{}}$0.7809^{++}$\\      (0.7623-0.7996)\end{tabular} & \begin{tabular}[c]{@{}c@{}}0.7820\\      (0.7618-0.8022)\end{tabular}  & \begin{tabular}[c]{@{}c@{}}$0.7749^*$\\      (0.7589-0.7908)\end{tabular} & \begin{tabular}[c]{@{}c@{}}$0.7587^*$\\      (0.7384-0.7790)\end{tabular} & \begin{tabular}[c]{@{}c@{}}$0.7772^*$\\      (0.7625-0.7918)\end{tabular} \\ \cline{2-9} 
	& KS    & \textbf{\begin{tabular}[c]{@{}c@{}}0.5087\\      (0.4951-0.5223)\end{tabular}} & \begin{tabular}[c]{@{}c@{}}$0.4231^+$\\      (0.4115-0.4346)\end{tabular}  & \begin{tabular}[c]{@{}c@{}}$0.4206^{++}$\\      (0.4055-0.4357)\end{tabular} & \begin{tabular}[c]{@{}c@{}}0.4630\\      (0.4466-0.4794)\end{tabular}  & \begin{tabular}[c]{@{}c@{}}$0.4100^*$\\      (0.3970-0.4229)\end{tabular} & \begin{tabular}[c]{@{}c@{}}$0.3785^*$\\      (0.3621-0.3949)\end{tabular} & \begin{tabular}[c]{@{}c@{}}$0.4225^*$\\      (0.4107-0.4344)\end{tabular} \\ \hline
	\multirow{2}{*}{11} & AUC   & \textbf{\begin{tabular}[c]{@{}c@{}}0.8022\\      (0.7897-0.8147)\end{tabular}} & \begin{tabular}[c]{@{}c@{}}0.7765\\      (0.7658-0.7871)\end{tabular}  & \begin{tabular}[c]{@{}c@{}}$0.7767^*$\\      (0.7628-0.7906)\end{tabular} & \begin{tabular}[c]{@{}c@{}}$0.7784^+$\\      (0.7634-0.7935)\end{tabular}  & \begin{tabular}[c]{@{}c@{}}$0.7679^*$\\      (0.7561-0.7798)\end{tabular} & \begin{tabular}[c]{@{}c@{}}$0.7582^*$\\      (0.7431-0.7734)\end{tabular} & \begin{tabular}[c]{@{}c@{}}$0.7736^*$\\      (0.7626-0.7845)\end{tabular} \\ \cline{2-9} 
	& KS    & \textbf{\begin{tabular}[c]{@{}c@{}}0.5082\\      (0.4983-0.5181)\end{tabular}} & \begin{tabular}[c]{@{}c@{}}0.4175\\      (0.4092-0.4259)\end{tabular}  & \begin{tabular}[c]{@{}c@{}}$0.4145^*$\\      (0.4035-0.4255)\end{tabular} & \begin{tabular}[c]{@{}c@{}}$0.4617^+$\\      (0.4498-0.4735)\end{tabular}  & \begin{tabular}[c]{@{}c@{}}$0.4014^*$\\      (0.3920-0.4108)\end{tabular} & \begin{tabular}[c]{@{}c@{}}$0.3963^*$\\      (0.3843-0.4082)\end{tabular} & \begin{tabular}[c]{@{}c@{}}$0.4183^*$\\      (0.4097-0.4270)\end{tabular} \\ \hline
	\multirow{2}{*}{12} & AUC   & \textbf{\begin{tabular}[c]{@{}c@{}}0.7988\\      (0.7899-0.8077)\end{tabular}} & \begin{tabular}[c]{@{}c@{}}$0.7716^*$\\      (0.7640-0.7791)\end{tabular}  & \begin{tabular}[c]{@{}c@{}}$0.7720^*$\\      (0.7622-0.7819)\end{tabular} & \begin{tabular}[c]{@{}c@{}}$0.7750^*$\\      (0.7643-0.7857)\end{tabular}  & \begin{tabular}[c]{@{}c@{}}$0.7637^*$\\      (0.7553-0.7722)\end{tabular} & \begin{tabular}[c]{@{}c@{}}$0.7506^*$\\      (0.7398-0.7613)\end{tabular} & \begin{tabular}[c]{@{}c@{}}$0.7693^*$\\      (0.7615-0.7771)\end{tabular} \\ \cline{2-9} 
	& KS    & \textbf{\begin{tabular}[c]{@{}c@{}}0.5053\\      (0.4985-0.5120)\end{tabular}} & \begin{tabular}[c]{@{}c@{}}$0.4121^*$\\      (0.4063-0.4178)\end{tabular}  & \begin{tabular}[c]{@{}c@{}}$0.4092^*$\\      (0.4017-0.4167)\end{tabular} & \begin{tabular}[c]{@{}c@{}}$0.4597^*$\\      (0.4516-0.4679)\end{tabular}  & \begin{tabular}[c]{@{}c@{}}$0.3961^*$\\      (0.3896-0.4025)\end{tabular} & \begin{tabular}[c]{@{}c@{}}$0.3870^*$\\      (0.3789-0.3952)\end{tabular} & \begin{tabular}[c]{@{}c@{}}$0.4126^*$\\      (0.4067-0.4186)\end{tabular} \\ \hline
\end{tabular}
}
\begin{tablenotes}
	\footnotesize
	\item[$*$] indicates p-value $ < 0.001$
	\item[$+$] indicates p-value $ < 0.01$
	\item[$++$] indicates p-value $ < 0.05$
\end{tablenotes}
\end{threeparttable}
\caption{AUC and KS indices of all candidate models (in terms of mean values and 95\% confidence intervals) for every repayment months. The highest score of each row is highlighted in boldface type.}
\label{tab:auc_ks_2}
\end{table}
}

\begin{figure}[!ht]
  \centering
    \includegraphics[width=0.8\textwidth]{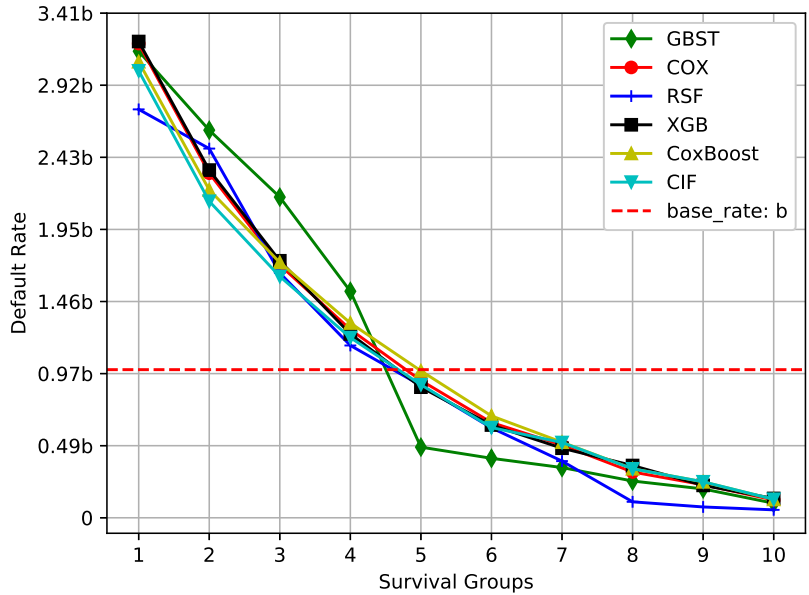}
   \caption{Default rate curves of different survival groups in the test dataset (March 2018) in 12th repayment month. $b$ is the same baseline as in Figure \ref{fig:default}, i.e., the default rate of the sample loans from the training dataset cumulated to the 12th month. Loans in the test dataset (March 2018) are sorted in ascending order by the survival probability predicted by all candidate models and are accordingly divided into 10 groups.}
   \label{fig:default3}
\end{figure}

Finally, we compare the predicted survival probabilities with the true default rate. For each model, the loans are sorted by the predicted survival probability of the 12th repayment month and are in ascending order divided into 10 survival groups with equal sample size. For each survival group predicted by each model, we calculate the true default rate of the 12th month within the group compared to the baseline $b$ that is the default rate of the loans from the training dataset cumulated to 12th month. The results are depicted in Figure \ref{fig:default3lc}. The baseline $b$ can also be approximately viewed as the average default rate of all 10 groups. It is interesting to find that when the survival groups are controlled, the GBST model has higher real default rates in the lower-ranked groups (group 1 -- 4, which are higher than the baseline) compared to the other models, while it has relatively lower real default rates in the higher-ranked groups, despite several exceptions from the RSF model. This finding proves that the GBST model is more capable of distinguishing the default loans from the others, which is in accordance with the previous results observed with the goodness-of-fit metrics.



\section{Conclusion and future work}

This paper presents a novel nonparametric ensemble tree model called gradient boosting survival tree (GBST) that extends the survival tree models with a gradient boosting algorithm. The survival tree ensemble is learned by optimizing the likelihood in an additive manner. As a test, the GBST model is applied to quantify credit risk on two large-scale real market datasets. The results on the both datasets show that the performance of GBST is better than that of previous models, measured by C-index, AUC and KS index of each time period. 

Like traditional scoring methods, the proposed gradient boosting survival tree (GBST) model is based on the attributes of the accounts (called \emph{features} in this paper) measured at the time of application. Yet, many of them change over time, especially when the modeling period is not short. To address the problem of changing attributes, \cite{djeundje2019dynamic} propose a dynamic parametric survival model with B-splines. In the future, we plan to incorporate time-varying attributes into the GBST model.

In the usual setting of survival analysis, as well as in the setting of GBST, one considers a binary state space, i.e., either the event of interest happens (state = 1) or not (state = -1). However, in many applications (see e.g., \citealp{lee2018deephit}), as in the problem of credit scoring, additional events may be of importance as well. For instance, an account with early repayment is viewed as a non-default account. However, with respect to profits, an early-repayment account tends to differ from an account with on-time repayments. This problem can be dealt with by extending the binary state space to the multi-state space, which is left in our future work.



\appendix
\begin{appendices}

\section{Maximum weighted quantile spilt algorithm}

The split algorithm for GBST stated in Algorithm \ref{alg:split} is a greedy algorithm exploring all possible split points, which can be extremely time-consuming when the value set of features is very large. In order to speed up, we propose here a maximum weighted quantile method by extending the method applied by \cite{chen2016xgboost}. Note that the objective in Eq.\ (13) can be rewrite as
\begin{align*}
 \tilde{\mathcal L}(f) &:= \sum_{j=1}^J \sum_{i \in N_j} \left( r_{i,j} f(\tau_j; \boldsymbol x_i) + \frac{1}{2} \sigma_{i,j} f^2(\tau_j; \boldsymbol x_i) \right) + \frac{\lambda}{2} \lVert w \rVert^2 \\
 &= \sum_{j=1}^J \sum_{i \in N_j} \left[ \frac{1}{2} \sigma_{i,j} \left( f(\tau_j; \boldsymbol x_i) - \left(- \frac{r_{i,j}}{\sigma_{i,j}}\right)  \right)^2 -\frac{1}{2} \frac{(r_{i,j})^2}{\sigma_{i,j}}  \right] + \frac{\lambda}{2} \lVert w \rVert^2. \\
\end{align*}
Hence, the objective can be viewed as a sum of the weighted squared loss with labels $- \frac{r_{i,j}}{\sigma_{i,j}}$ and weights $\sigma_{i,j}$.

At time $\tau_j$, we consider a multi-set $N^k_j := \{(x_{1,k}, \sigma_{1,j}), (x_{2,k}, \sigma_{2,j}), \ldots, (x_{n_j,k}, \sigma_{n_j,j}) \}$ satisfying that $x_{1, k} \leq x_{2, k} \leq \ldots \leq x_{n_j,k}$, where $n_j$ denotes the total size of samples that survive longer than $\tau_{j-1}$.  In other words, the survived samples are sorted by the value of its $k$th feature. Then we define a rank function $g^j_k : \mathbb{R} \rightarrow [0, \infty)$  at time $\tau_j$ such that
\begin{align*}
 g^j_k(z) := \frac{\sum_{(x, \sigma) \in N^k_j, x < z} \sigma}{\sum_{(x,\sigma) \in N^k_j} \sigma}.
\end{align*}
With the rank function, the candidate split points $\{s_{k,1}, s_{k,2},...,s_{k,q} \}$ are given by
\begin{align*}
& s_{k,1} = \min_i x_{i,k}, \\
& s_{k,q} = \max_i x_{i,k}, \\
&\max_{j_1 \in \{1,2,\ldots,{J(t_d) \wedge J(t_{d+1}) \wedge J} \}} | g^{j_1}_k(s_{k, d}) - g^{j_1}_k(s_{k, d+1})| < \varepsilon, \\
&\max_{j_2 \in \{1,2,\ldots, {J(t_d) \wedge J(t_{i}) \wedge J} \}} | g^{j_2}_k(s_{k, d}) - g^{j_2}_k(x_{i,k})| \geq \varepsilon, \quad \forall x_{i,k} > s_{k, d+1}, i \in N_{j_2},  \\
&\max_{j_3 \in \{1,2,..., {J(t_d) \wedge J(t_{i}) \wedge J}\}} | g^{j_3}_k(s_{k, d}) - g^{j_3}_k(x_{i,k})| < \varepsilon, \quad \forall s_{k, d} > x_{i,k} > s_{k, d+1}, i \in N_{j_3},
\end{align*}
where $\varepsilon$ is the approximation factor and one can prove that the maximum size of split points is $J \cdot \lceil \frac{1}{\varepsilon} \rceil$ \citep{chen2016xgboost}. The algorithm for generating split points is summarized in Algorithm \ref{alg:split_weight}. Accordingly, the revised split algorithm with maximum weighted quantile is demonstrated in Algorithm \ref{alg:split_new}.

\begin{algorithm}[H]
  \caption{Search for split points with maximum weighted quantile method.}
  \label{alg:split_weight}
  \begin{algorithmic}[1]
    \Require
     $I$: Set of individuals.
    \For{$j=1$ to $J$}
    \For{$k=1$ to $K$}
    \State $N^k_j = \{(x_{1,k}, \sigma_{1,j}), (x_{2,k}, \sigma_{2,j}), ..., (x_{n_j,k}, \sigma_{n_j,j}) \}$
    \State $s^j_{k,1} \leftarrow \min_i x_{i,k}, i \in N_j$
    \State $s^j_{k,q} \leftarrow \max_i x_{i,k}, i \in N_j$
    \State $S^j = \{ s^j_{k,1} \}$
    \State $s^j_{k,d} \leftarrow s^j_{k,1}$
    \For{$x_{i,k}$ in sorted $N^k_j$ and $x_{i,k} \geq s^j_{kd}$}
    \If{$| g^{j}_k(s_{k, d}) - g^{j}_k(x_{i,k})| < \varepsilon$ and $| g^{j}_k(s_{k, d}) - g^{j}_k(x_{i+1,k})| \geq \varepsilon$}
    \State $ s^j_{k,d} \leftarrow x_{i,k}$
    \State $S^j += \{ s^j_{k,d} \}$
    \EndIf
    \EndFor
    \State $S^j += \{ s^j_{k,q} \}$
    \EndFor
    \EndFor
    \Ensure
      $S = S^1 \cup S^2 \cup ... \cup S^J$
  \end{algorithmic}
\end{algorithm}

\begin{algorithm}[H]
  \caption{Maximum weighted quantile split algorithm for GBST.}
  \label{alg:split_new}
  \begin{algorithmic}[1]
    \Require
     $I$: Set of individuals in current node; $\left\{ \boldsymbol x_i \in \mathbb R^n \right\}$: the features of individuals; $N_j, j = 1, 2, \ldots, J$ defined in Eq. (9)
    \For{$k=1$ to $K$}
    \State Get the sorted set $S_k$ of split points with the maximum weighted quantile method
    \EndFor
    \For{$k=1$ to $K$}
    \For{$j=1$ to $J$}
    \State $W_j\leftarrow \sum_{i \in N_j \cap I} r_{i, j}$
    \State $V_j\leftarrow \sum_{i \in N_j \cap I} \sigma_{i, j}$
    \EndFor
    \For{$s$ in $S_k$}
    \For{$j=1$ to $J$}
    \State $W_{jL}\leftarrow \sum_{i \in \{ i| x_i \in N_j, x_{i,k} \leq s \}} r_{i, j}$
    \State $V_{jL} \leftarrow \sum_{i \in \{ i| x_i \in N_j, x_{i,k} \leq s \}} \sigma_{i, j}$
    \State $W_{jR}=W_j-W_{jL}$
    \State $V_{jR}=V_j-V_{jL}$
    \EndFor
    \State $score \leftarrow \max(score, \sum_j[\frac{W_{jL}^2}{V_{jL}+\lambda}+\frac{W_{jR}^2}{V_{jR}+\lambda}-\frac{W_{j}^2}{V_{j}+\lambda}]$)
    \EndFor
    \EndFor
    \Ensure
      Split with max score, splitting schema
  \end{algorithmic}
\end{algorithm}

\section{Parallel execution}
Another trick to speed up the GBST algorithm is to apply parallel computing. For boosting methods, trees must be generated sequentially since the successive tree depends on the training and prediction of the previous tree. Nevertheless, in the process of training a single tree, the problem of finding the best split points (see Algorithm \ref{alg:split_weight}) can be solved in parallel. Before training each tree, we can pre-sort the features and reuse the sorting result in subsequent iterations, which can significantly reduce the amount of calculation and improve algorithm performance.

\section{Links of open-source packages}
\label{app:links}
\setlength{\tabcolsep}{7mm}{
\begin{table}[!ht]
\centering
\resizebox{\textwidth}{!}{%
\begin{tabular}{c|l}
\hline
{\bf Model}& {\bf Link Address} \\
\hline
GBST & https://github.com/360jinrong/GBST\\
COX & https://lifelines.readthedocs.io/en/latest/fitters/regression/CoxPHFitter.html\\
RSF & https://scikit-survival.readthedocs.io/en/latest/api/generated/sksurv.ensemble.RandomSurvivalForest.html\\
CoxBoost & https://cran.r-project.org/web/packages/CoxBoost/index.html \\
CIF & https://cran.r-project.org/web/packages/party/party.pdf \\
GBMCI &  https://github.com/uci-cbcl/GBMCI  \\
DeepHit &  https://github.com/havakv/pycox  \\
\hline
\end{tabular}}
\label{table:model}
\end{table}
}

\end{appendices}

\end{document}